\documentclass{article}

\usepackage{microtype}
\usepackage{graphicx}
\usepackage{subcaption}
\usepackage{booktabs} 

\usepackage{hyperref}

\usepackage[preprint]{icml2026}

\usepackage{amsmath}
\usepackage{amssymb}
\usepackage{mathtools}
\usepackage{amsthm}
\usepackage{tabularx}
\usepackage{tcolorbox}
\usepackage{array}
\usepackage{booktabs}

\usepackage[utf8]{inputenc} 
\usepackage[T1]{fontenc}    
\usepackage{hyperref}       
\usepackage{url}            
\usepackage{booktabs}       
\usepackage{amsfonts}       
\usepackage{nicefrac}       
\usepackage{microtype}      
\usepackage{xcolor}         
\usepackage{soul}
\usepackage{graphicx}
\usepackage{multirow} 
\usepackage[capitalize,noabbrev,nameinlink]{cleveref}
\usepackage{tikz}
\usepackage{adjustbox}
\usetikzlibrary{positioning,calc}

\theoremstyle{plain}

\theoremstyle{definition}

\theoremstyle{remark}

\usepackage[textsize=tiny]{todonotes}

\icmltitlerunning{Submission for International Conference on Machine Learning (ICML 2026)}

\begin{document}

\twocolumn[
  \icmltitle{Toward Trustworthy Evaluation of Sustainability Rating Methodologies: \\A Human–AI Collaborative Framework for Benchmark Dataset Construction}



  \icmlsetsymbol{equal}{*}

 \begin{icmlauthorlist}
    \icmlauthor{Xiaoran Cai}{equal,yyy}
    \icmlauthor{Wang Yang}{equal,case}
    \icmlauthor{Xiyu Ren}{hkust}
    \icmlauthor{Chekun Law}{nvidia}
    \icmlauthor{Rohit Sharma}{informatica}
    \icmlauthor{Peng Qi}{uniphore}
\end{icmlauthorlist}

\icmlaffiliation{yyy}{Columbia University, USA}
\icmlaffiliation{case}{Case Western Reserve University, USA}
\icmlaffiliation{hkust}{Hong Kong University of Science and Technology, Hong Kong}
\icmlaffiliation{nvidia}{NVIDIA, USA}
\icmlaffiliation{informatica}{Informatica Inc., USA}
\icmlaffiliation{uniphore}{Uniphore, USA}

\icmlcorrespondingauthor{Xiaoran Cai}{xc2422@columbia.edu}
\icmlcorrespondingauthor{Wang Yang}{wxy320@case.edu}

  \icmlkeywords{Machine Learning, ICML}

  \vskip 0.3in
]




\printAffiliationsAndNotice{}

\begin{abstract} 
Sustainability or ESG rating agencies use company disclosures and external data to produce scores or ratings that assess the environmental, social, and governance performance of a company. However, sustainability ratings across agencies for a single company vary widely, limiting their comparability, credibility, and relevance to decision-making. To harmonize the rating results, we propose adopting a universal human-AI collaboration framework to generate trustworthy benchmark datasets for evaluating sustainability rating methodologies. The framework comprises two complementary parts: \textbf{STRIDE} (Sustainability Trust Rating \& Integrity Data Equation) provides principled criteria and a scoring system that guide the construction of firm-level benchmark datasets using large language models (LLMs), and \textbf{SR-Delta}, a discrepancy-analysis procedural framework that surfaces insights for potential adjustments. The framework enables scalable and comparable assessment of sustainability rating methodologies. We call on the broader AI community to adopt AI-powered approaches to strengthen and advance sustainability rating methodologies that support and enforce urgent sustainability agendas.

\end{abstract}

\section{Introduction} 
Sustainability has become a central concern for governments, investors, consumers, and corporations, as it increasingly shapes long-term economic resilience, social well-being, and environmental stability \cite{HuberComstock2017ESG, eccles2014impact}. In response, companies embed sustainability into their operations and strategic planning and disclose performance through sustainability reports that include quantitative and qualitative indicators such as carbon emissions, labor practices, and corporate governance structures \cite{odinaka2025sustainability}. 

However, companies differ in their disclosure strategies, limiting comparability. As a result, sustainability ratings were introduced as intermediaries that translate non-comparable information across companies into decision-relevant insights \cite{lu2024regulating}. Sustainability ratings serve as a multipurpose informational infrastructure, guiding investors' capital allocation, supporting corporate performance improvement, informing regulators and policymakers, and enabling public stakeholders such as consumers and NGOs to assess corporate responsibility and engage in advocacy \cite{oecd2025behind}.

As sustainability ratings have gained prominence, their proliferation has introduced a new issue: rating results for a company are inconsistent across different agencies. Previous research showed that the correlation among ESG~\footnote{Please note that the terms "ESG" and "sustainability" are used interchangeably in this paper. This is because the input documents are not uniformly labeled as "ESG reports" or "sustainability reports". Companies adopt different naming conventions depending on their strategic objectives, reporting systems, and the standards they follow.} ratings ranges from 0.38 to 0.71 due to methodological differences in scope (38\%), weighting (6\%), and measurement (56\%) \cite{berg2022aggregate}, creating confusion for stakeholders and ambiguity in the market \cite{SLGI_NoTwoESGRatings}. For example, in 2023, Sustainalytics rated Adani's governance as high risk ("moderate"), while Morgan Stanley Capital International ("MSCI") maintained an "A" rating for Adani Green Energy. Following the Hindenburg report, Adani's shares plunged, wiping £100 billion off the company's market value \cite {KerberWilkes2023}. Such evidence highlights structural biases and commercial influences in sustainability rating methodologies \cite{Zhang2024, berg2025esg}. Consequently, these issues undermine the credibility and practical application of these ratings \cite{berg2022aggregate} and further raise questions about the financial benefits of sustainability \cite{AdamopoulouGkizori2025}.

Sustainability ratings or overall sustainability remain underexplored within the AI research community in two dimensions: technology focus areas and scope of study. First, previous related work in sustainability focuses mainly on extracting unstructured data from ESG reports \cite{zou2023esgreveal,sun2024information}, on annotation \cite{wu2025aiannotator}, or on vertical applications of AI such as decarbonization and smart cities \cite{toderas2025ai_sustainability}. Limited efforts have focused on sustainability ratings as a primary research object. Second, the input data have a specific scope by region or sector \cite{sun2024information}, limiting their comparability. For metric selection, most studies adopt a single international framework or regional guideline \cite{zou2023esgreveal}, often neglecting firm-specific and industry-specific metrics essential for contextualized evaluation. Other work further narrows its scope by focusing exclusively on a single category of sustainability metrics, such as environmental indicators \cite{angioni2024investigating}, thereby failing to capture the multidimensional nature of sustainability performance. 

\textbf{Our Position}: A universal human–AI collaboration framework that guides the generation of benchmark datasets and produces scores to assess their readiness should be adopted to evaluate sustainability rating methodologies.

From an impact standpoint, there are more than 600 ESG or sustainability ratings as of 2018 \cite{wong2020ratetheraters}. In the UK alone, more than 150 sustainability rating providers operate in the market, and regulatory intervention could save firms an estimated £578 million over the next decade\cite{FT_ESGRatings2025}.   

From a technology standpoint, AI and LLMs have demonstrated capabilities in information extraction, reasoning over unstructured disclosures, and synthesizing qualitative and quantitative information at scale \cite{bronzini2024glitter}. A multi-agent system has been used to extract KPIs from financial reports \cite{choi2025structuring}. These capabilities empower scalable development of standardized insight-level evaluation frameworks for the evaluation of sustainability rating methodologies.  
 
To support this position, this paper makes the following key contributions:

\begin{itemize}
    \item Propose STRIDE: The Sustainability Trust Rating \& Integrity Data Equation (STRIDE) is a universal framework with key consideration elements for generating firm-level benchmark datasets with a scoring system.
    \item Propose SR-Delta: The SR-Delta framework uses STRIDE guided benchmark datasets for discrepancy analysis to generate improvement insights. 
    \item Provide alternative perspectives: Review other possible approaches, identify associated barriers, and present our responses.
\end{itemize}

The frameworks are developed based on insights from previous research and our empirical experience in a case study. No optimal framework exists, and continuous improvement is necessary over time. We provide an illustrative case study as an initial step with the intention of stimulating feedback and iterative refinement over time.  

\section{Key Challenges in AI-Based Sustainability Rating Benchmarking}

\textbf{Challenge 1: Inconsistent Sustainability Practices} 

Unlike financial disclosure, sustainability disclosure lacks a standardized set of guidelines that companies can consistently reference \cite{berg2022aggregate}. Heterogeneous standards and guidelines at the global, national, local, industry, and firm levels result in inconsistent sustainability disclosure practices \cite{stedman2025esgframeworks}. Even more than that, regulations and policies at the same level can be fragmented \cite{park2024extraterritoriality}. For example, countries exhibit heterogeneous sustainability priorities, regulatory frameworks, and reporting languages. Consequently, the governance, alignment, and standardization of sustainability disclosure remain an unresolved challenge in sustainability assessment and benchmarking\cite{berg2022aggregate}.

As for rating agencies, they typically adopt one or more reporting standards to design their rating methodologies. For example, MSCI uses 150 policy metrics, 20 performance metrics, and more than 100 governance key metrics based on its own business model \cite {MSCI2024RatingsMethodology}, while Sustainalytics uses 180 metrics across 138 sub-industries\cite{sustainalytics_impact_reporting}. 

\textbf{Challenge 2: Limited High-quality Input Data} 

 Sustainability ratings rely on corporate sustainability disclosures as data input \cite{Windolph2011}, yet the quality of these inputs remains a concern. Incentives for sustainability disclosure are regulatory in nature \cite{DeVilliers2020Trends}. Mandatory sustainability disclosure is limited to a small number of jurisdictions (e.g, the EU under the CSRD), with voluntary reporting prevailing elsewhere \cite{Hatter2025Transatlantic}. As a result, voluntary disclosure companies often approach sustainability disclosure as a branding or public relations exercise in the absence of clear regulatory or economic incentives, emphasizing narrative and storytelling over standardized measurement \cite{Hassani2024DiscourseEmissions}. Therefore, the input data may be misaligned with real-world conditions and exhibit quality deficiencies. Even for companies with the capacity and intention to engage in sustainability disclosure, the reporting information infrastructures, such as data systems and software, remain uneven and underdeveloped, limiting the collection and integration of reliable sustainability data \cite{Troshani2024SustainabilityInfrastructure}.  

\textbf{Challenge 3: Lack of an End-to-end Trustworthy Document Acquisition Layer}

Sustainability reports consist of complex, long, unstructured data, including text, tables, and figures. There is a lack of reliable agentic pipelines for report retrieval, human annotation, provenance tracking across corporate web properties, and agile feedback loops for content drift. The current LLM model for understanding ESG-related knowledge achieves a measurable level of accuracy, typically around 55-70\%, and the results indicate room for improvement \cite{he-etal-2025-esgenius}. Human-annotated data is often introduced to compensate for these limitations. Yet, there is no consensus that ground-truth datasets reliably improve model performance, and prior work highlights risks arising from subjective judgments, inconsistent annotation guidelines, and varying annotator expertise \cite{zajac2023ground,foody2024ground,zhang2020automatic}. These challenges are further amplified by the scale and fragmentation of sustainability-related data on the web, with a rapidly growing stream of external news and civil society reporting. Sustainability concepts themselves also evolve and are interpreted inconsistently, necessitating agile, end-to-end systems rather than static models or datasets \cite{berg2022aggregate}.

\begin{figure*}[ht]
    \centering
    \includegraphics[width=\linewidth]{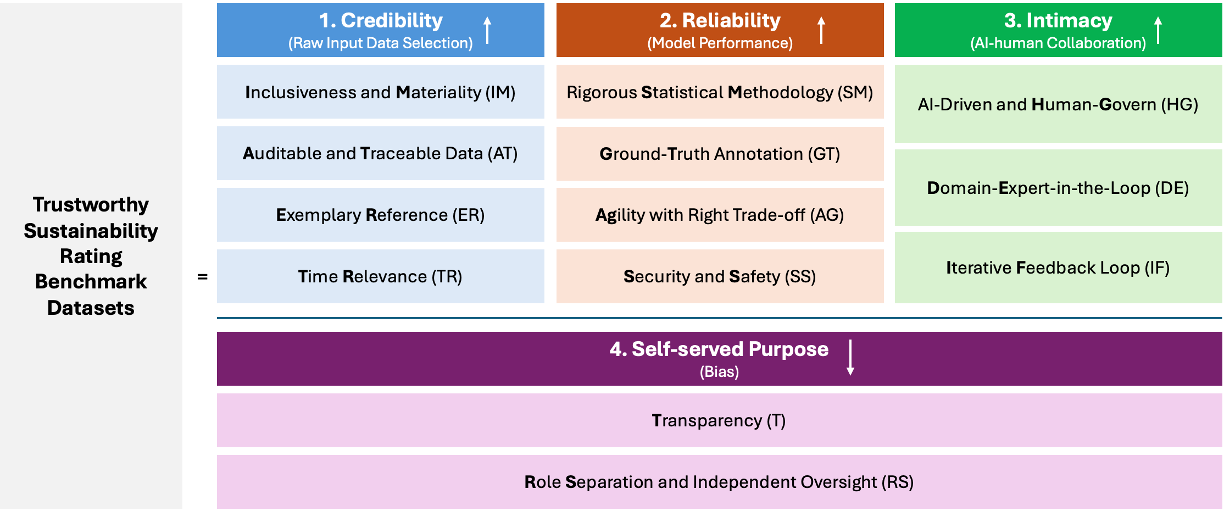}
    \caption{STRIDE framework for trustworthy sustainability rating benchmark datasets. Trust is modeled as a function of credibility, reliability, and human–AI intimacy (positive contributors), and self-serving purpose (negative contributor).}
    \label{fig:trust_equation2}
\end{figure*}
\begin{figure*}[ht]
\centering
\begin{tcolorbox}[
  title={STRIDE: Sustainability Trust Rating and
Integrity Data Equation},
  colback=white,
  colframe=black,
  fonttitle=\bfseries,
  boxrule=0.6pt,      
  arc=1pt,            
  left=4pt,
  right=4pt,
  top=2pt,
  bottom=2pt,
  before skip=4pt,    
  after skip=4pt
]

\begin{equation}
\label{eq:trust}
\begin{aligned}
\tau(x)
&=
\sigma\!\left(
\alpha_C C(x)
+
\alpha_R R(x)
+
\alpha_I I(x)
-
\alpha_S S(x)
\right),&
\sigma(z)=
\frac{1}{1+\exp(-z)} .
\end{aligned}
\end{equation}

\begin{equation}
\label{eq:credibility}
C(x)
=
w_{\mathrm{IM}}^{C}\,\mathrm{IM}(x)
+
w_{\mathrm{AT}}^{C}\,\mathrm{AT}(x)
+
w_{\mathrm{ER}}^{C}\,\mathrm{ER}(x)
+
w_{\mathrm{TR}}^{C}\,\mathrm{TR}(x).
\end{equation}

\begin{equation}
\label{eq:reliability}
R(x)
=
w_{\mathrm{SM}}^{R}\,\mathrm{SM}(x)
+
w_{\mathrm{GT}}^{R}\,\mathrm{GT}(x)
+
w_{\mathrm{AG}}^{R}\,\mathrm{AG}(x)
+
w_{\mathrm{SS}}^{R}\,\mathrm{SS}(x).
\end{equation}

\begin{equation}
\label{eq:intimacy}
I(x)
=
w_{\mathrm{HG}}^{I}\,\mathrm{HG}(x).
+
w_{\mathrm{DE}}^{I}\,\mathrm{DE}(x)
+
w_{\mathrm{IF}}^{I}\,\mathrm{IF}(x) 
\end{equation}

\begin{equation}
\label{eq:selfserved}
S(x)
=
w_{\mathrm{T}}^{S}\,\mathrm{T}(x)
+
w_{\mathrm{RS}}^{S}\,\mathrm{RS}(x).
\end{equation}

\vspace{0.4em}
\noindent\textbf{Parameters.}
$\tau(x)\in(0,1)$ denotes the overall human--machine trust score.
$C(x)$, $R(x)$, $I(x)$, and $S(x)$ represent credibility, reliability,
human--AI intimacy, and self-served purpose, respectively.
$\alpha_\bullet \ge 0$ control cross-dimension importance.
Weights $w_{\bullet}^{(\cdot)} \ge 0$ form convex combinations within each component.

\vspace{0.4em}
\noindent\textbf{Sub-metrics.}
$\mathrm{IM}$ (Inclusiveness \& Materiality),
$\mathrm{AT}$ (Auditable \& Traceable Data),
$\mathrm{ER}$ (Exemplary Reference),
$\mathrm{TR}$ (Time Relevance),
$\mathrm{SM}$ (Rigorous Statistical Methodology),
$\mathrm{GT}$ (Ground-Truth Annotation),
$\mathrm{AG}$ (Agility with Right Trade-off),
$\mathrm{SS}$ (Security \& Safety),
$\mathrm{HG}$ (AI-Driven and Human-Governed),
$\mathrm{DE}$ (Domain-Expert-in-the-Loop),
$\mathrm{IF}$ (Iterative Feedback Loop),
$\mathrm{T}$ (Transparency),
$\mathrm{RS}$ (Role Separation and Independent Oversight).

\end{tcolorbox}
\vspace{-6pt}
\caption{STRIDE trust formulation and component decomposition.}
\label{fig:trust_box}
\end{figure*}

\section{STRIDE: \underline{S}ustainability \underline{T}rust \underline{R}ating and \underline{I}ntegrity  \underline{D}ata \underline{E}quation}

The Trust Equation, introduced in 2000, conceptualizes human-to-human trust as a function of four components: credibility, reliability, and intimacy, which enhance trust, and self-orientation, which diminishes it \cite{green2000trusted}. Now, the Trust Equation can be extended to human–machine interaction research, where it helps to understand and design trust in machine-generated outputs, including decision-support systems and AI-enabled tools. 

Previous studies show that perceptions of credibility, reliability, and intent alignment play a central role in whether users accept, rely on, or override machine recommendations \cite{larasati2025human}. Existing research on trustworthy AI in industrial contexts informs the design of a structured trustworthiness data framework that spans data preparation, algorithm design, system development, deployment, and operational workflows \cite{li2023trustworthy}. In this work, we formulate the trust equation to account for the domain-specific nature of sustainability ratings and the collaboration between LLMs and human experts.  

\paragraph{Human-Machine Trust Equation.} The overview of the framework about our trust equation is shown in \Cref{fig:trust_equation2}. We model human-machine trust as a latent scalar $\tau(x) \in (0,1)$, defined as a sigmoid function of a weighted linear combination of four components (as shown in \Cref{fig:trust_box}): credibility $C(x)$, reliability $R(x)$, human--AI intimacy $I(x)$, and self-serving purpose $S(x)$. The sigmoid function $\sigma(x)$ ensures that the trust score is bounded between 0 and 1 and exhibits a diminishing sensitivity to extreme values of the latent trust signal.

Following the Trust Equation, AI-generated benchmark data reflect industry best practices rather than average or minimal compliance \cite{amelzadeh2018why}. The appropriate depth and breadth of data and systems are designed to generate comparable and actionable insights \cite{oecd2025behind}. The details of the function and equations are provided in \Cref{app:equations}.


\subsection{Credibility}

Credibility provides principled guidance for selecting raw input data. It guides and sets criteria for the sufficiency of data coverage and the verifiability of data sources. It indicates data quality. Credibility includes four aspects: \textit{inclusiveness and materiality $\mathrm{IM}(x)$, auditable and traceable data $\mathrm{AT}(x)$, exemplary reference $\mathrm{ER}(x)$, and time relevance $\mathrm{TR}(x)$}.



\textbf {Inclusiveness and Materiality ($\mathrm{IM}$)} capture whether a sustainability rating benchmark dataset framework includes relevant issues across sectors \cite{eccles2012need} and regions while properly focusing on metrics that matter for impact, risk, and decision-making. 

Two input sources are used: sustainability reports that contain data values, and standards and principles that define metrics. First, for sustainability reports, it is critical to be representative across sectors and regions to ensure that the benchmark dataset is generalizable as a credible reference \cite{dutoit2024sustainability}. Second, for metrics, there are five standard layers: global, national, local, industry-specific, and company-specific. Uniform metrics may fail to capture industry, geographic, and contextual differences in corporate behavior and priorities \cite{shi2025esg_divergence}. Companies also disclose tailored quantitative data for their operations, making "one-size-fits-all" metrics difficult to maintain materiality \cite{liu2022quantitative}. Third, beyond internal and disclosure data, it is essential to incorporate metrics that capture external data, such as sentiment and narratives in local news, which provide independent signals of real-world impacts, stakeholder perceptions, and emerging sustainability risks 
\cite{Hassan_2019_FirmLevelRisk}.

$\mathrm{IM}(x)$ can be defined as a function of geographic coverage, industry coverage,
standard-layer coverage, and the incorporation of external datasets:
\[
\mathrm{IM}(x)
=
f\!\left(
\mathcal{C}_{\mathrm{ctry}}(x),
\mathcal{I}_{\mathrm{ind}}(x),
\mathcal{S}(x),
\mathcal{E}(x)
\right).
\]



\textbf {Auditable and Traceable Data ($\mathrm{AT}$)} refer to selected input data that are supported by verifiable evidence internally and externally. The source of the data can be traced throughout the data lifecycle, thereby mitigating the challenges posed by subjective and narrative-driven sustainability disclosures. 

For data audibility, the absence of auditable pipelines and robust governance controls can compromise the credibility of datasets and, therefore, mislead model performance \cite {luo2024automate}. Traceability enables transparency, accountability, and reproducibility by allowing users to audit input values and data transformations throughout the analytical pipeline \cite{buneman2001provenance,bouthillier2021survey}. Traceability also supports risk management and governance by enabling organizations to identify, diagnose, and correct errors or biases introduced at specific stages of data collection and processing \cite{pineau2021improving}.

$\mathrm{AT}(x)$ can be defined as a function of auditability and traceability
across the evidence set $\mathcal{E}(x)$:
\[
\mathrm{AT}(x)
=
f\!\left(
\mathrm{aud}(e),
\mathrm{tr}(e),
\mathcal{E}(x)
\right).
\]



\textbf {Exemplary Reference} requires data sources from organizations with established reporting practices so that the data reflect credible operations with practical value for insights. 

First, benchmark datasets often draw on longitudinal disclosures from high-governance companies such as FTSE 350 constituents \cite{Ferjancic2024TextualESG}. Second, comparative sustainability analyses typically rely on high-quality disclosures from well-governed firms as reference data to ensure robustness and confidence level~\cite{Meng2025ComparativeCSR}. Systematic reviews of ESG indicators emphasize the need to compare with consistent high-assurance corporate disclosures to support comparability \cite{daCunha2025ESGFramework}.

$\mathrm{ER}(x)$ can be defined as a function of the confidence level $\sigma$ associated with the sustainability disclosure of a company and the number of recognitions it receives in the sustainability field:
\[
\mathrm{ER}(x)
=
f\!\left(
\sigma,
n_{\mathrm{rec}}(x)
\right).
\]



\textbf {Time Relevance} refers to the extent to which data reflect the most relevant and current conditions, practices, and contextual realities at the time of analysis.

Unlike annual sustainability reports, rating agencies' timelines do not necessarily align with companies' periodic reporting cycles, and rating outcomes can change in response to data or methodological updates. For example, MSCI ESG Research distinguishes between full rating actions and interim data updates. While comprehensive ESG Ratings and Industry-Adjusted Scores are recalculated only during formal rating reviews, updates to individual data points can automatically affect relevant scores, often within a week, without an immediate analyst-led reassessment \cite{MSCI2025FAQ}.  

$\mathrm{TR}(x)$ can be defined as a function of the overlap evaluation time and a temporal decay rate for time relevancy:
\[
\mathrm{TR}(x)
=
f\!\left(
\Delta t(x),
\lambda
\right).
\]



\subsection{Reliability}
Reliability addresses the rigor and robustness of model performance. It guides and sets criteria for modeling processes and output validation. It signals model quality. Reliability includes four dimensions: \textit{rigorous sampling methodology $\mathrm{SM}(x)$, ground-truth annotation $\mathrm{GT}(x)$, agility with appropriate trade-offs $\mathrm{AG}(x)$, and security and safety $\mathrm{SS}(x)$}. 


\textbf {Rigorous Statistical Methodology} refers to the use of statistical techniques that ensure sampling results are sufficient, representative, and robust to noise.

 First, the sample size should be chosen to achieve saturation, beyond which additional data do not result in statistically significant performance improvements \cite{Ott2022BenchmarkSaturation}. Second, a selected sample dataset is representative if it has minimal distributional deviation from the population \cite{petersen2016representative}. Beyond examining surface-level data properties, it is equally important to consider the substantive characteristics of the subject matter being studied \cite{hornung2023evaluating}. For example, when constructing benchmark datasets from sustainability reports, sampling should reflect diversity across companies and industries, as well as variation in sustainability themes and disclosure practices.

$\mathrm{SM}(x)$ can be defined as a function of the sample saturation threshold and a deviation-based representativeness measure:
\[
\mathrm{SM}(x)
=
f\!\left(
n_{\mathrm{sat}},
\sigma_p(x)
\right).
\]


\textbf {Ground-Truth Annotation} refers to the process of creating reference labels or judgments by humans that serve as baselines against which models, systems, or benchmarks are evaluated. 

First, prior research shows that models trained with human-verified subsets outperform models trained on fully automated labels, even when the human-labeled portion is relatively small (10–30\%) \cite{zhang2021human}. Second, while aggregated non-expert annotations can approximate expert labels for well-defined and low-interpretation tasks, expert annotations differ qualitatively by encoding deeper domain knowledge, greater conceptual consistency, and more reliable handling of ambiguous and long-tail cases such as sustainability reports \cite {snow2008cheap}. Third, annotation disagreement can be signals \cite{aroyo2015truth} and improve downstream model robustness \cite{uma2021annotator}. 

$\mathrm{GT}(x)$ can be defined as a function of the proportion of
human-validated annotations, expert
utilization signals, and
deviation-based agreement terms capturing human-machine annotation alignment and machine-machine annotation alignment: 
\[
\mathrm{GT}(x)
=
f\!\left(
p_{\mathrm{human}}(x),
u(x),
\sigma_{HM}(x),
\sigma_{MM}(x)
\right).
\]



\textbf {Agility with Right Trade-off} refers to a model’s ability to adapt and improve in response to changing conditions while maintaining performance.

In rapidly evolving domains such as sustainability and AI governance, benchmark systems must remain adaptable without compromising methodological integrity \cite{Hardt2025EmergingBench}. Agility in benchmark design enables iterative refinement, incorporation of newly emerging standards, and responsiveness to changing practices to maintain and improve model performance \cite{DynamicBenchmark2503}. At the same time, explicit articulation of trade-offs, such as the use of partial automation, proxy variables, or sampled annotations, is essential to ensure transparency, interpretability, and justified methodological choices.

$\mathrm{AG}(x)$ can be defined as a function of the marginal accuracy gain per change of the model:
\[
\mathrm{AG}(x)
=
f\!\left(
\mathrm{Acc}(x),
\mathrm p(x)
\right).
\]



\textbf {Security and Safety} refers to the ability of AI to be falsely confident and generate adversarial examples.

The risk of sustainability rating benchmarks is the generation of biased or misleading insights that distort decision-making \cite{berg2022aggregate}, particularly when users treat benchmark output as proxies for the underlying real-world sustainability performance. 

$\mathrm{SS}(x)$ can be defined as a function of the number of rows that contain harmful information
$N_{\mathrm{harm}}(x)$ and the total number of extracted rows
$N_{\mathrm{total}}(x)$:
\[
\mathrm{SS}(x)
=
f\!\left(
N_{\mathrm{harm}}(x),
N_{\mathrm{total}}(x)
\right).
\]



\subsection{Intimacy}
Intimacy refers to the degree to which AI systems, domain experts, and governance stakeholders work together \cite {shneiderman2020human}. It guides and sets criteria for human-machine collaboration. It represents the extent to which the design of the system enables structured interaction between human experts and AI models, allowing humans to guide, interpret, challenge, and refine the results of the model \cite{bansal2021does}. Intimacy includes three dimensions: \textit{AI-driven and human-govern $\mathrm{HG}(x)$, domain-expert-in-the-Loop $\mathrm{DE}(x)$, iterative feedback loop $\mathrm{IF}(x)$}. 


\textbf {AI-Driven and Human-Govern} describes systems in which AI performs tasks, while humans retain oversight, decision-making authority, and accountability for system design and outputs.

Given the inherently context-dependent and normative nature of sustainability, humans define rating objectives and materiality boundaries, validate model outputs, and intervene when models exhibit error, bias, or drift \cite{vaccaro2024humanai}. AI systems enable the efficient processing of large, complex, and unstructured datasets, while human governance ensures contextual understanding, ethical oversight, and alignment with societal values\cite{floridi2018aigovernance}. 

$\mathrm{HG}(x)$ can be defined as a function of the number of human intervention cases and the total number of generated cases. 

\[
\mathrm{HG}(x)
=
f\!\left(
n_{\mathrm{intervene}}(x),
n_{\mathrm{cases}}(x)
\right).
\]

\textbf {Domain-Expert-in-the-Loop} refers to the active involvement of subject-matter experts in the design, annotation, validation, and governance of data and model outputs for model robustness.

Domain-expert-in-the-loop has been proven to enhance the robustness of AI and ML techniques. This approach improves the performance of knowledge graphs by improving semantic accuracy, data quality, and schema design throughout the graphs' life-cycle \cite{Bikaun2024CleanGraph}. Previous research shows that expert participation reduces noise in automated extraction, strengthens ontology alignment, and provides higher-quality training and evaluation data for knowledge graph learning models. These contributions lead to more reliable inference, improved downstream task performance, and greater robustness in dynamic and domain-specific settings \cite{Zhang2019HumanKG}. 

$\mathrm{DE}(x)$ can be defined as a function of the number of domain-expert involvement across an AI pipeline, the number of pipeline stages with expert participation, the total number of domain experts, and pipeline stages within the scope of the system:

\[
\mathrm{DE}(x)
=
f\!\left(
\mathcal{E}(x),
\mathcal{S}(x),
N_{\mathcal{E}},
N_{\mathcal{S}}
\right).
\]


\textbf {Iterative Feedback Loop} refers to a structured process in which system outputs are continuously evaluated, corrected, and refined through repeated cycles of human and/or automated feedback.

An iterative feedback loop is critical for improving system performance, reliability, and alignment over time. In benchmark construction and Human–AI collaboration, feedback loops enable the identification of systematic errors, evolving edge cases, and shifts in domain expectations \cite{Raji2020Closing}. Prior research demonstrates that iterative feedback is essential for maintaining adaptability, particularly in complex, real-world domains where static benchmarks quickly become outdated \cite{Raji2020Closing}. Studies on interactive machine learning also show that trial-and-error–based iterative evaluation and collaboration between humans and AI systems constitute a core methodology for testing and improving human-AI collaboration \cite{mosqueira2023humanloop}. 

$\mathrm{IF}(x)$ can be defined as a function of the number of human and/or automated feedback at iteration that leads to explicit correction and improvement in the accuracy rate. 

\[
\mathrm{IF}(x)
=
f\!\left(
\mathcal{I}(x),
\Delta x
\right).
\]



\subsection{Self-served Purpose}
Self-served Purpose addresses the extent to which potential biases influence benchmarking outcomes. It guides and sets criteria for disclosure transparency and governance. It represents transparency in the disclosure of assumptions, constraints, and potential conflicts of interest \cite{suresh2021framework} as well as the independence of the evaluation procedures. The self-served purpose includes two dimensions: \textit{transparency $\mathrm{T}(x)$, and role separation and independent evaluation $\mathrm{RS}(x)$}. 



\textbf {Transparency} refers to the explicit documentation and communication of the underlying assumptions, methodological choices, data constraints, and known limitations that shape the design and output of the system \cite{Kroll2017Accountable}. 

Explicitly documenting assumptions about data coverage, proxies, and model scope along with known limitations such as data gaps or uncertainty helps prevent misuse and overgeneralization \cite{brundage2020toward}. Previous work in responsible AI and evaluation methodology emphasizes transparency as a prerequisite for accountability, reproducibility, and trust, particularly when benchmarks inform decisions or policy discussions \cite{inel2023collect}. 

 $\mathrm{T}(x)$ can be defined as a function of the assumptions required to build the benchmark dataset unit and the number of explicitly disclosed assumptions.

\[
\mathrm{T}(x)
=
f\!\left(
\mathcal{A}_{\mathrm{dis}}(x),
\mathcal{A}_{\mathrm{req}}(x),
\Delta x
\right).
\]


\textbf {Role Separation and Independent Evaluation} refer to the structural separation of responsibilities across system design, implementation, and evaluation.

Role separation operates across three dimensions: AI agent roles, human roles, and human–AI collaboration roles. Research on algorithmic governance and accountability across human roles highlights the importance of delineating responsibilities among developers, testers, and oversight bodies to mitigate risks in algorithmic systems \cite{Cobbe2021Reviewable, Cech2021AlgorithmicAccountability, Blake2024AlgorithmicAccountability}. Research on algorithmic governance and AI system design emphasizes that AI agents should have different levels of data access to support role separation. When agents are allowed to produce and assess results, the risks of circular validation, misaligned rewards, and unchecked bias increase \cite{Kroll2017Accountable, Raji2020Closing, Cobbe2021Reviewable}.

 $\mathrm{RS}(x)$ can be separated as the subset of relations whose responsibilities and handovers should be separated into units and a function of the set of all potential relationships among the roles.

\[
\mathrm{RS}(x)
=
f\!\left(
\mathcal{E}_{\mathrm{sep}}(x),
\mathcal{E}^\star
\right).
\]
 

\section {SR-Delta: \underline{S}ustainability \underline{R}ating \underline{D}elta for Methodology Improvement}

For both existing and newly developed sustainability rating methodologies (Step 1), we gather original results based on the specified methodologies (Step 2) using agencies’ proprietary data. In parallel, the same methodology is applied to a STRIDE-guided benchmark data set that standardizes data quality, coverage, and interpretability (Step 3). Holding the rating logic constant across both pipelines ensures that any differences in outcomes arise from data construction rather than methodological design. Comparing the resulting rating outputs (Steps 4 and 5) enables systematic discrepancy analysis (Step 6), which identifies and further analyzes data-driven sources of rating divergence. 

The STRIDE framework does not prescribe corrective actions; rather, it surfaces signals and warnings for deeper analysis. To validate our position, we applied the STRIDE and SR-Delta frameworks to Luxshare Precision Industry Co. ("Luxshare"), a Fortune~500 electronics manufacturer.

We constructed a STRIDE-guided benchmark dataset for Luxshare using its 2024 sustainability report. Based on the proposed scoring parameters, the dataset achieved an overall STRIDE score of $\tau = 0.56$. We used the datasets to go through the MSCI sustainability rating methodology to compare the results  (\textquotedblleft BB\textquotedblright) for the same reporting period.

The incumbent executive pay rating is affected by disclosure ambiguity, whereas the chemical safety rating shows systematic over-estimation. The STRIDE-guided approach exposes these inconsistencies, illustrating how the human–AI collaborative benchmarking framework can serve as a rigorous evaluation layer for conventional sustainability ratings. Please see an end-to-end case study in \Cref{app:casestudy}.

\begin{figure*}
    \centering
    \includegraphics[width=1\linewidth]{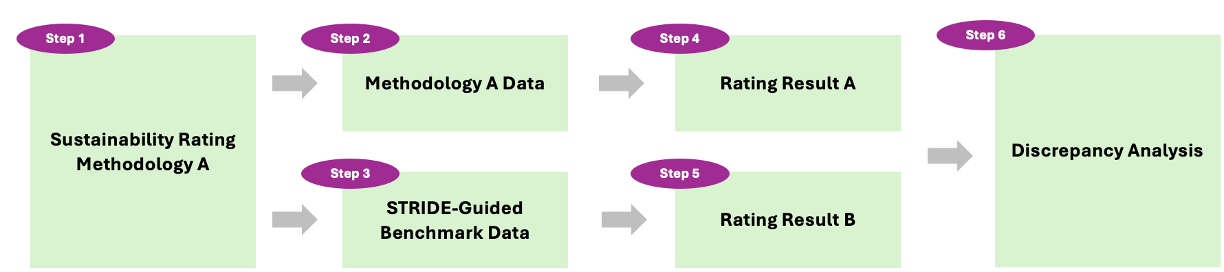}
    \caption{Overview of the STRIDE-guided discrepancy analysis framework. Rating outcomes generated using an existing sustainability rating methodology (A) are compared against ratings produced using STRIDE-guided benchmark data to identify systematic divergences.}
    \label{fig:process_graph.png}
\end{figure*}

\section{Alternative Views} 

\textbf{View 1: There are more efficient ways of solving the problem rather than using an AI-driven benchmark dataset.} Early sustainability efforts were largely driven by political movements and public initiatives \cite{Park2022EvolutionSustainability}. Regulatory convergence (e.g., mandatory disclosure standards) can reduce rating divergence at the source by improving data consistency between firms, making AI-driven benchmark datasets a secondary rather than a primary solution \cite{Kong2025IndustryESG}. Local community movements are also drivers of sustainability norms and action, especially for environmental change \cite{Ostrom2010Polycentric}. 

\textbf{Response:} The benchmark dataset methodology serves as an accelerator rather than a replacement. Benchmark datasets have played an essential role in advancing machine learning by enabling standardized evaluation, comparability, and rapid iteration. Early benchmarks such as MNIST (1998) accelerated algorithmic progress \cite{LeCun1998MNIST}. ImageNet (2009), with its unprecedented scale and challenge-based evaluation, exposed the limitations of existing computer vision methods and directly incentivized the development of new architectures, most notably deep convolutional neural networks \cite{Deng2009ImageNet}. 

Time also matters when it comes to the sustainability schedule. The IPCC AR6 Synthesis Report warns that global warming will likely exceed the critical 1.5~$^\circ$C threshold in the early 2030s, beyond which climate risks rise non-linearly, drastic and irreversible impacts increase\cite{IPCC_AR6_Synthesis_2023, IPCC_SR15_2018}. Sustainability ratings and reports are essential inputs for measuring progress \cite{OECD_2021_SustainabilityMetrics}. Having a standardized framework for benchmark datasets is therefore essential to ensure consistency, comparability, and accountability across sustainability assessments, enabling timely, evidence-based decision-making aligned with climate targets \cite{berg2022aggregate}.

\textbf{View 2: The application of AI can introduce biases in benchmark datasets.} AI can hallucinate \cite{OpenAI_2023_GPT4}. AI models can generate confidence without unfounded facts. When AI-generated outputs are reused iteratively to update benchmarks, such hallucinations can propagate and harden to apparent ground truth, creating self-reinforcing distortions that undermine the reliability and the use of decisions of sustainability assessments \cite{Shumailov_2024_ModelCollapse}.

\textbf{Response:} The framework does not aim for complete accuracy or transparency; rather, it is designed to generate a benchmark dataset for insights and support critical thinking. Achieving deeper and more reliable insights requires human–AI collaboration, where AI systems surface patterns, anomalies, and alternative perspectives, while human judgment provides contextual understanding, ethical reasoning, and domain expertise \cite{Amershi_2019_HumanAI}. Through this side-by-side interaction, a form of intimacy between human and machine emerges, which enables iterative sense-making, challenges overconfidence in automated outputs, and strengthens interpretive rigor rather than replacing human responsibility \cite {Shneiderman_2020_HumanCenteredAI}.

\textbf{View 3: Stakeholders resist the changes.} Implementing an AI-driven or unified benchmarking framework may face resistance from firms, regulators, and investors anchored to incumbent methodologies, driven by concerns over transparency, interpretability, and potential reputation or re-rating risks \cite{Recht2019ImageNetGeneralize}.

\textbf{Response:} Such resistance is a well-documented feature of methodological and technological change, particularly in high-stakes decision environments. Research on human-centered and responsible AI emphasizes that adoption is more likely when systems are positioned as decision-support tools rather than automated arbiters, preserving human judgment, interpretability, and accountability. By framing the framework as an augmentation instrument based on insight rather than a replacement for existing processes, it can facilitate gradual adoption, trust calibration, and stakeholder acceptance, aligning with best practices in human–AI collaboration and change governance \cite{Amershi_2019_HumanAI, Shneiderman_2020_HumanCenteredAI}. A data-driven science-based solution has been more effective in driving change. 

\section{Conclusion}


To address discrepancies in sustainability ratings, we propose an AI–human collaborative, firm-level benchmarking framework, \textbf{STRIDE}, together with a discrepancy analysis procedure, \textbf{SR-Delta}, which enables systematic identification of divergence sources and improvement opportunities in rating methodologies.

We hope that this position catalyzes community engagement toward an open, auditable, and continuously updated benchmark infrastructure for the evaluation of sustainability rating methodologies. We call on the broader AI community to recognize the urgency of sustainability challenges and to adopt AI-powered approaches to harmonize rating discrepancies, allowing for more accurate measurement of sustainability progress.

\newpage

\bibliography{example_paper}

@article{HuberComstock2017ESG,
  title   = {ESG Reports and Ratings: What They Are, Why They Matter},
  author  = {Huber, Betty Moy and Comstock, Michael},
  journal = {Harvard Law School Forum on Corporate Governance},
  year    = {2017},
  month   = jul,
  day     = {27},
  note    = {Posted by Davis Polk \& Wardwell LLP}
}

@unpublished{Park2022EvolutionSustainability,
  title     = {The Evolution of Sustainability Concerns over Business Activities: From Local to Cross-National to Global},
  author    = {Park, Junghoon and Cuervo-Cazurra, Alvaro and Montiel, Ivan},
  year      = {2022},
  note      = {Manuscript EE003, November 27, 2022},
  institution = {Baruch College, City University of New York and Northeastern University}
}

@article{Kong2025IndustryESG,
  title   = {Can Industry-Specific Information Disclosure Guidelines Alleviate Corporate ESG Diverence? Evidence from Chinese Listed Companies},
  author  = {Kong, Linghui and Chen, Rongquan and Huang, Xinyu and Wang, Fan},
  journal = {International Review of Financial Analysis},
  year    = {2025},
  pages   = {104529},
  doi     = {10.1016/j.irfa.2025.104529},
  url     = {https://doi.org/10.1016/j.irfa.2025.104529}
}

@article{Ostrom2010Polycentric,
  title   = {Polycentric Systems for Coping with Collective Action and Global Environmental Change},
  author  = {Ostrom, Elinor},
  journal = {Global Environmental Change},
  volume  = {20},
  number  = {4},
  pages   = {550--557},
  year    = {2010},
  doi     = {10.1016/j.gloenvcha.2010.07.004}
}

@article{LeCun1998MNIST,
  title   = {Gradient-Based Learning Applied to Document Recognition},
  author  = {LeCun, Yann and Bottou, L{\'e}on and Bengio, Yoshua and Haffner, Patrick},
  journal = {Proceedings of the IEEE},
  volume  = {86},
  number  = {11},
  pages   = {2278--2324},
  year    = {1998},
  doi     = {10.1109/5.726791}
}

@inproceedings{Deng2009ImageNet,
  title     = {ImageNet: A Large-Scale Hierarchical Image Database},
  author    = {Deng, Jia and Dong, Wei and Socher, Richard and Li, Li-Jia and Li, Kai and Fei-Fei, Li},
  booktitle = {Proceedings of the IEEE Conference on Computer Vision and Pattern Recognition (CVPR)},
  year      = {2009},
  pages     = {248--255},
  doi       = {10.1109/CVPR.2009.5206848}
}

@techreport{IPCC_AR6_Synthesis_2023,
  author       = {{Intergovernmental Panel on Climate Change}},
  title        = {Climate Change 2023: Synthesis Report},
  institution  = {IPCC},
  year         = {2023},
  address      = {Geneva, Switzerland},
  note         = {Contribution of Working Groups I, II and III to the Sixth Assessment Report}
}

@techreport{OECD_2021_SustainabilityMetrics,
  author       = {{Organisation for Economic Co-operation and Development}},
  title        = {Sustainability Metrics: Measurement for Policy Action},
  institution  = {OECD},
  year         = {2021},
  address      = {Paris, France}
}

@techreport{IPCC_SR15_2018,
  author       = {{Intergovernmental Panel on Climate Change}},
  title        = {Global Warming of 1.5{$^\circ$}C},
  institution  = {IPCC},
  year         = {2018},
  address      = {Geneva, Switzerland},
  note         = {Special Report on the impacts of global warming of 1.5{$^\circ$}C}
}

@article{eccles2014impact,
  title={The Impact of Corporate Sustainability on Organizational Processes and Performance},
  author={Eccles, Robert G. and Ioannou, Ioannis and Serafeim, George},
  journal={Management Science},
  volume={60},
  number={11},
  pages={2835--2857},
  year={2014},
  publisher={INFORMS}
}

@article{lu2024regulating,
  title={Regulating ESG rating firms as the gatekeepers for sustainable finance},
  author={Lu, Longjie},
  journal={Capital Markets Law Journal},
  volume={19},
  number={2},
  pages={184--206},
  year={2024},
  publisher={Oxford University Press}
}

@article{odinaka2025sustainability,
  title={Sustainability Practices in Fortune 500 Companies and Their Impact on Business Practices: A Multiple Case Study Analysis},
  author={Odinaka, Nnadozie and Wash-Anigboro, Oghenetega},
  journal={Journal of Energy Research and Reviews},
  volume={17},
  number={7},
  pages={17--26},
  year={2025}
}

@article{berg2022aggregate,
  title   = {Aggregate Confusion: The Divergence of ESG Ratings},
  author  = {Berg, Florian and K{\"o}lbel, Julian F. and Rigobon, Roberto},
  journal = {Review of Finance},
  volume  = {26},
  number  = {6},
  pages   = {1315--1344},
  year    = {2022},
  doi     = {10.1093/rof/rfac033}
}

@article{Hassan_2019_FirmLevelRisk,
  author  = {Hassan, Tarek A. and Hollander, Stephan and van Lent, Laurence and Tahoun, Ahmed},
  title   = {Firm-Level Political Risk: Measurement and Effects},
  journal = {Quarterly Journal of Economics},
  volume  = {134},
  number  = {4},
  pages   = {2135--2202},
  year    = {2019}
}

@article{Shumailov_2024_ModelCollapse,
  author  = {Shumailov, Ilia and Shumaylov, Ziyaad and Zhao, Yiren and Gal, Yarin and Papernot, Nicolas and Anderson, Ross},
  title   = {AI Models Collapse When Trained on Recursively Generated Data},
  journal = {Nature Machine Intelligence},
  year    = {2024},
  doi     = {10.1038/s42256-024-00813-4}
}

@article{Amershi_2019_HumanAI,
  author  = {Amershi, Saleema and Weld, Dan and Vorvoreanu, Mihaela and Fourney, Adam and Nushi, Besmira and Collisson, Penny and Suh, Jina and Iqbal, Shamsi and Bennett, Paul N. and Inkpen, Kori and Teevan, Jaime and Kikin-Gil, Rosa and Horvitz, Eric},
  title   = {Guidelines for Human-AI Interaction},
  journal = {Proceedings of the ACM on Human-Computer Interaction},
  volume  = {3},
  number  = {CSCW},
  pages   = {1--24},
  year    = {2019}
}

@inproceedings{Recht2019ImageNetGeneralize,
  title     = {Do ImageNet Classifiers Generalize to ImageNet?},
  author    = {Recht, Benjamin and Roelofs, Rebecca and Schmidt, Ludwig and Shankar, Vaishaal},
  booktitle = {Proceedings of the 36th International Conference on Machine Learning},
  year      = {2019},
  url       = {https://arxiv.org/abs/1902.10811}
}

@article{petersen2016representative,
  title   = {Representative sampling for reliable data analysis: Theory of Sampling},
  author  = {Petersen, Lars and Minkkinen, Pentti and Esbensen, Kim H.},
  journal = {Chemometrics and Intelligent Laboratory Systems},
  volume  = {158},
  pages   = {1--20},
  year    = {2016}
}

@article{aroyo2015truth,
  title   = {Truth Is a Lie: Crowd Truth and the Seven Myths of Human Annotation},
  author  = {Aroyo, Lora and Welty, Chris},
  journal = {AI Magazine},
  volume  = {36},
  number  = {1},
  pages   = {15--24},
  year    = {2015},
  doi     = {10.1609/aimag.v36i1.2564}
}

@inproceedings{uma2021annotator,
  title     = {Annotator Disagreement, Label Ambiguity, and Model Performance},
  author    = {Uma, Alexandra and Foroutan, Negar and Miller, Tim and Gurevych, Iryna},
  booktitle = {Proceedings of the 2021 Conference on Empirical Methods in Natural Language Processing},
  pages     = {7560--7575},
  year      = {2021}
}

@inproceedings{zhang2021human,
  title     = {Human-in-the-Loop Learning for Robust Neural Networks},
  author    = {Zhang, Yifan and others},
  booktitle = {Advances in Neural Information Processing Systems},
  year      = {2021}
}

@article{li2023trustworthy,
  title     = {Trustworthy AI: From Principles to Practices},
  author    = {Li, Bo and Qi, Peng and Liu, Bo and Di, Shuai and Liu, Jingen and Pei, Jiquan and Yi, Jinfeng and Zhou, Bowen},
  journal   = {ACM Computing Surveys},
  volume    = {55},
  number    = {9},
  pages     = {1--46},
  year      = {2023},
  publisher = {ACM}
}

@misc{OpenAI_2023_GPT4,
  author = {{OpenAI}},
  title  = {GPT-4 Technical Report},
  year   = {2023},
  note   = {Acknowledges hallucination as a core limitation of large language models}
}

@article{Shneiderman_2020_HumanCenteredAI,
  author  = {Shneiderman, Ben},
  title   = {Human-Centered Artificial Intelligence: Reliable, Safe \& Trustworthy},
  journal = {International Journal of Human–Computer Interaction},
  volume  = {36},
  number  = {6},
  pages   = {495--504},
  year    = {2020}
}

@misc{SLGI_NoTwoESGRatings,
  title        = {No two ESG ratings are the same},
  author       = {{Sun Life Global Investments Institutional}},
  year         = {2023},
  month        = {January 30},
  howpublished = {\url{https://www.slgiinstitutional.com/en/news-and-insights/No-two-ESG-ratings-are-the-same/}},
  note         = {Accessed: 2025-12-22},
}

@misc{FT_ESGRatings2025,
  title        = {UK Watchdog Seeks to ‘Clean Up’ Conflicts of Interest in ESG Ratings},
  author       = {{Financial Times}},
  year         = {2025},
  month        = {December 1},
  howpublished = {\url{https://www.ft.com/content/93bbbc83-0799-41fe-a3f1-48792d8834c8}},
  note         = {Accessed: 2025-12-22},
}

@article{KerberWilkes2023,
  author = {Ross Kerber and Tommy Wilkes},
  title = {Sustainalytics downgrades three Adani companies' governance scores},
  journal = {Reuters},
  year = {2023},
  month = {February 9},
  url = {https://www.reuters.com/article/adani-group-sustainalytics-idUSL4N34X2XE}
}

@article{Zhang2024,
  author = {Liandong Zhang and others},
  title = {Do commercial ties influence ESG ratings? Evidence on conflicts of interest in ESG rating agencies},
  journal = {Journal of Accounting Research (as discussed via Wiley press release)},
  year = {2024},
  note = {Commercial relationships and acquisitions can bias ESG ratings, increasing scores for clients of parent firms},
  url = {https://onlinelibrary.wiley.com/doi/10.1111/1475-679X.12582}
}

@article{AdamopoulouGkizori2025,
  author = {Evgenia Adamopoulou and Konstantina Gkizori},
  title = {The Impact of ESG Investments on Market Capitalization},
  journal = {Preprints.org},
  year = {2025},
  doi = {10.20944/preprints202507.0415.v1},
  url = {https://www.preprints.org/manuscript/202507.0415}
}

@article{zou2023esgreveal,
  title={ESGReveal: An LLM-based Approach for Extracting Structured Data from ESG Reports},
  author={Zou, Yi and Shi, Mengying and Chen, Zhongjie and Deng, Zhu and Lei, ZongXiong and Zeng, Zihan and Yang, Shiming and Tong, HongXiang and Xiao, Lei and Zhou, Wenwen},
  journal={arXiv preprint arXiv:2312.17264},
  year={2023},
  doi={10.48550/arXiv.2312.17264},
  primaryClass={cs.CL}
}

@inproceedings{sun2024information,
  title={Information extraction: Unstructured to structured for esg reports},
  author={Sun, Zounachuan and Satapathy, Ranjan and Guo, Daixue and Li, Bo and Liu, Xinyuan and Zhang, Yangchen and Tan, Cheng-Ann and Shirota Filho, Ricardo and Goh, Rick Siow Mong},
  booktitle={2024 IEEE International Conference on Data Mining Workshops (ICDMW)},
  pages={487--495},
  year={2024},
  organization={IEEE}
}

@article{wu2025aiannotator,
  title={The AI Annotator: Large Language Models’ Potential in Scoring Sustainability Reports},
  author={Wu, Yue and Hu, Peng and Wang, Derek D.},
  journal={Systems},
  volume={13},
  number={10},
  pages={899},
  year={2025},
  publisher={MDPI},
  doi={10.3390/systems13100899},
  url={https://doi.org/10.3390/systems13100899}
}

@inproceedings{he-etal-2025-esgenius,
  title     = {ESGenius: Benchmarking LLMs on Environmental, Social, and Governance (ESG) and Sustainability Knowledge},
  author    = {He, Chaoyue and Zhou, Xin and Wu, Yi and Yu, Xinjia and Zhang, Yan and Zhang, Lei and Wang, Di and Lyu, Shengfei and Xu, Hong and Xiaoqiao, Wang and Liu, Wei and Miao, Chunyan},
  booktitle = {Proceedings of the 2025 Conference on Empirical Methods in Natural Language Processing},
  month     = nov,
  year      = {2025},
  address   = {Suzhou, China},
  publisher = {Association for Computational Linguistics},
  pages     = {14623--14664},
  doi       = {10.18653/v1/2025.emnlp-main.739},
  url       = {https://aclanthology.org/2025.emnlp-main.739/}
}

@article{Bikaun2024CleanGraph,
  title   = {CleanGraph: Human-in-the-Loop Knowledge Graph Refinement and Completion},
  author  = {Bikaun, Tyler and Stewart, Michael and Liu, Wei},
  journal = {None},
  year    = {2024},
  note    = {arXiv preprint},
  url     = {https://arxiv.org/abs/2405.03932}
}

@article{Zhang2019HumanKG,
  title   = {Human-in-the-Loop Knowledge Graph Construction},
  author  = {Zhang, Ziqi and Li, Juanzi and others},
  journal = {IEEE Intelligent Systems},
  volume  = {34},
  number  = {6},
  pages   = {44--52},
  year    = {2019}
}

@article{Raji2020Closing,
  title   = {Closing the AI Accountability Gap: Defining an End-to-End Framework for Internal Algorithmic Auditing},
  author  = {Raji, Inioluwa Deborah and Smart, Andrew and White, Rebecca and Mitchell, Margaret and Gebru, Timnit and others},
  journal = {Proceedings of the ACM Conference on Fairness, Accountability, and Transparency (FAccT)},
  year    = {2020}
}

@article{hornung2023evaluating,
  title   = {Evaluating Machine Learning Models in Non-Standard Settings: An Overview and New Findings},
  author  = {Hornung, Roman and Nalenz, Malte and Schneider, Lennart and Bender, Andreas and Bothmann, Ludwig and Bischl, Bernd and Augustin, Thomas and Boulesteix, Anne-Laure},
  journal = {arXiv preprint arXiv:2310.15108},
  year    = {2023},
  doi     = {10.48550/arXiv.2310.15108},
  eprint  = {2310.15108},
  archivePrefix = {arXiv},
  primaryClass  = {stat.ML}
}

@article{Ott2022BenchmarkSaturation,
  title   = {Mapping global dynamics of benchmark creation and saturation in artificial intelligence},
  author  = {Ott, Simon and Barbosa-Silva, Adriano},
  journal = {Scientific Data},
  year    = {2022},
  note    = {Discusses benchmark saturation patterns where additional data yield limited performance improvements},
  url     = {https://www.ncbi.nlm.nih.gov/pmc/articles/PMC9649641/}
}

@misc{Hardt2025EmergingBench,
  author       = {Hardt, Moritz},
  title        = {The Emerging Science of Machine Learning Benchmarks},
  year         = {2025},
  howpublished = {SIAM News; manuscript available online at mlbenchmarks.org},
  note         = {Discusses limitations of static benchmarks and the need for evolving evaluation frameworks},
  url          = {https://mlbenchmarks.org/}
}

@article{DynamicBenchmark2503,
  author    = {B. Guan},
  title     = {Is Your Benchmark (Still) Useful? Dynamic Benchmarking through Input Transformation},
  journal   = {arXiv preprint},
  year      = {2025},
  volume    = {abs/2503.06643},
  url       = {https://arxiv.org/abs/2503.06643}
}

@article{floridi2018aigovernance,
  title={AI4People—An ethical framework for a good AI society},
  author={Floridi, Luciano and Cowls, Josh and Beltrametti, Monica and others},
  journal={Minds and Machines},
  year={2018},
  volume={28},
  number={4},
  pages={689--707},
  doi={10.1007/s11023-018-9482-5}
}

@misc{sustainalytics_impact_reporting,
  author       = {{Sustainalytics}},
  title        = {Impact Metrics: Analyze and Report on the Impact of Your Portfolio},
  howpublished = {\url{https://www.sustainalytics.com/impact-reporting}},
  year         = {n.d.},
  note         = {Accessed January 4, 2026}
}

@article{vaccaro2024humanai,
  title     = {When Combinations of Humans and AI Are Useful: A Systematic Review and Meta-Analysis},
  author    = {Vaccaro, Michelle and Almaatouq, Abdullah and Malone, Thomas W.},
  journal   = {Nature Human Behaviour},
  year      = {2024},
  doi       = {10.1038/s41562-024-02024-1}
}

@article{mosqueira2023humanloop,
  title     = {Human-in-the-loop Machine Learning: A State of the Art},
  author    = {Mosqueira-Rey, E. and Hernández-Pereira, E. and Alonso-Ríos, D. and Bobes-Bascarán, J. and Fernández-Leal, A.},
  journal   = {Artificial Intelligence Review},
  volume    = {56},
  pages     = {3005--3054},
  year      = {2023},
  doi       = {10.1007/s10462-022-10246-w}
}

@book{green2000trusted,
  title={The Trusted Advisor},
  author={Green, Charles H. and Maister, David H. and Galford, Robert M.},
  year={2000},
  publisher={Free Press},
  address={New York, NY}
}

@article{larasati2025human,
  title={Human and AI Trust: Trust Attitude Measurement Instrument},
  author={Larasati, Retno and De Liddo, Anna and Motta, Enrico},
  journal={arXiv preprint arXiv:2510.21535},
  year={2025},
  url={https://arxiv.org/abs/2510.21535}
}

@article{toderas2025ai_sustainability,
  title={Artificial Intelligence for Sustainability: A Systematic Review and Critical Analysis of AI Applications, Challenges, and Future Directions},
  author={Toderas, Mihaela},
  journal={Sustainability},
  volume={17},
  number={17},
  pages={8049},
  year={2025},
  publisher={MDPI},
  doi={10.3390/su17178049},
  url={https://doi.org/10.3390/su17178049}
}

@article{dutoit2024sustainability,
  title   = {Thirty Years of Sustainability Reporting: Insights, Gaps and an Agenda for Future Research Through a Systematic Literature Review},
  author  = {Du Toit, Elmarie},
  journal = {Sustainability},
  volume  = {16},
  number  = {23},
  pages   = {10750},
  year    = {2024},
  publisher = {MDPI},
  doi     = {10.3390/su162310750}
}

@article{luo2024automate,
  title={The More You Automate, the Less You See: Hidden Pitfalls of AI Scientist Systems},
  author={Luo, Ziming and Kasirzadeh, Atoosa and Shah, Nihar B.},
  journal={arXiv preprint},
  year={2024},
  institution={Carnegie Mellon University},
  note={arXiv preprint}
}

@article{eccles2012need,
  title={The need for sector-specific materiality and sustainability reporting standards},
  author={Eccles, Robert G. and Krzus, Michael P. and Rogers, Jean and Serafeim, George},
  journal={Journal of Applied Corporate Finance},
  volume={24},
  number={2},
  pages={65--71},
  year={2012},
  doi={10.1111/j.1745-6622.2012.00380.x}
}

@article{bouthillier2021survey,
  title={Survey of Machine Learning Experimental Methodology},
  author={Bouthillier, Xavier and Laurent, C{\'e}sar and Vincent, Pascal},
  journal={Journal of Machine Learning Research},
  volume={22},
  number={202},
  pages={1--76},
  year={2021}
}

@article{pineau2021improving,
  title={Improving Reproducibility in Machine Learning Research},
  author={Pineau, Joelle and Vincent-Lamarre, Philippe and Sinha, Koustuv and Larivi{\`e}re, Vincent and Beygelzimer, Alina and d'Alch{\'e}-Buc, Florence and Fox, Emily and Larochelle, Hugo},
  journal={Journal of Machine Learning Research},
  volume={22},
  number={164},
  pages={1--20},
  year={2021}
}

@techreport{MSCI2024RatingsMethodology,
  author       = {{MSCI ESG Research LLC}},
  title        = {MSCI ESG Ratings Methodology: Process},
  institution  = {MSCI Inc.},
  year         = {2024},
  month        = apr,
  url          = {https://www.msci.com/documents/1296102/34424357/MSCI+ESG+Ratings+Methodology+-+Process.pdf},
  note         = {Accessed: 2025-01-30}
}

@article{Ferjancic2024TextualESG,
  title   = {Textual Analysis of Corporate Sustainability Reporting and Corporate ESG Scores},
  author  = {Ferjan{\v{c}}i{\v{c}}, Ur{\v{s}}a and
             Ichev, Riste and
             Lon{\v{c}}arski, Igor and
             Montariol, Syrielle and
             Pelicon, Andra{\v{z}} and
             Pollak, Senja and
             Sitar {\v{S}}u{\v{s}}tar, Katarina and
             Toman, Ale{\v{s}} and
             Valentin{\v{c}}i{\v{c}}, Aljo{\v{s}}a and
             {\v{Z}}nidar{\v{s}}i{\v{c}}, Martin},
  journal = {Journal of Business Research},
  year    = {2024},
  volume  = {175},
  pages   = {114610},
  issn    = {0148-2963},
  doi     = {10.1016/j.jbusres.2024.114610}
}

@article{daCunha2025ESGFramework,
  author  = {da Cunha, {\'I}caro Guilherme F{\'e}lix and
             Policarpo, Renata Veloso Santos and
             de Oliveira, Paula Cristina Senra and
             Abdala, Etienne Cardoso and
             Rebelatto, Daisy Aparecida do Nascimento},
  title   = {A Systematic Review of ESG Indicators and Corporate Performance: Proposal for a Conceptual Framework},
  journal = {Future Business Journal},
  year    = {2025},
  volume  = {11},
  number  = {1},
  pages   = {106},
  doi     = {10.1186/s43093-025-00539-1},
  issn    = {2314-7210}
}

@misc{MSCI2025FAQ,
  author       = {{MSCI ESG Research LLC}},
  title        = {General FAQs for Corporate Issuers},
  year         = {2025},
  month        = jun,
  publisher    = {MSCI},
  url          = {https://www.msci.com/documents/1296102/10259127/FAQ-For-Corporate-Issuers.pdf},
  note         = {Accessed: 2025-01-31}
}

@article{Kroll2017Accountable,
  title   = {Accountable Algorithms},
  author  = {Kroll, Joshua A. and Huey, Joanna and Barocas, Solon and Felten, Edward W. and Reidenberg, Joel R. and Robinson, David G. and Yu, Harlan},
  journal = {University of Pennsylvania Law Review},
  volume  = {165},
  pages   = {633--705},
  year    = {2017}
}

@article{brundage2020toward,
  title={Toward Trustworthy AI Development: Mechanisms for Supporting Verifiable Claims},
  author={Brundage, Miles and Avin, Shahar and Wang, Jasmine and others},
  journal={arXiv preprint arXiv:2004.07213},
  year={2020}
}

@inproceedings{inel2023collect,
  title={Collect, Measure, Repeat: Reliability Factors for Responsible AI Data Collection},
  author={Inel, Oana and Draws, Tim and Aroyo, Lora},
  booktitle={Proceedings of the AAAI Conference on Human Computation and Crowdsourcing (HCOMP)},
  year={2023},
  note={arXiv:2308.12885}
}

@article{Cech2021AlgorithmicAccountability,
  author  = {Cech, Filip and others},
  title   = {Mechanisms for Algorithmic Accountability Through the Lens of Actor and Forum Agency},
  journal = {Technology in Society},
  year    = {2021},
  note    = {Discusses the need for stakeholder agency and governance to support transparency and accountability}
}

@article{Cobbe2021Reviewable,
  title   = {Reviewable Automated Decision-Making: A Framework for Accountable Algorithmic Systems},
  author  = {Cobbe, Jennifer and Lee, Michelle Seng Ah and Singh, Jatinder},
  journal = {Law, Innovation and Technology},
  volume  = {13},
  number  = {1},
  pages   = {93--126},
  year    = {2021}
}

@misc{Blake2024AlgorithmicAccountability,
  author       = {Blake, Harrison},
  title        = {Algorithmic Accountability: Establishing Frameworks for Transparency and Responsibility in AI-driven Decisions},
  year         = {2024},
  note         = {Explores accountability frameworks that support separation of roles and oversight in AI systems},
  howpublished = {Preprint / Research article}
}

@article{Meng2025ComparativeCSR,
  author  = {Meng, Qiao and Knapp, Daniel and Brecht, Leo and Eckert, Roland},
  title   = {A Comparative Analysis of Corporate Sustainability Reporting: A Multi-Method Approach to China and the United States},
  journal = {Sustainability},
  year    = {2025},
  volume  = {17},
  number  = {22},
  pages   = {10315},
  doi     = {10.3390/su172210315},
  issn    = {2071-1050}
}

@inproceedings{buneman2001provenance,
  title={Why and where: A characterization of data provenance},
  author={Buneman, Peter and Khanna, Sanjeev and Tan, Wang-Chiew},
  booktitle={International Conference on Database Theory},
  pages={316--330},
  year={2001},
  organization={Springer}
}

@article{berg2025esg,
  title={ESG Ratings: A Compass without Direction},
  author={Berg, Florian and Kölbel, Julian F. and Rigobon, Roberto},
  year={2025},
  journal={SSRN Electronic Journal},
  url={https://download.ssrn.com/2025/10/31/4536239.pdf?abstractId=4536239},
  note={Accessed December 25, 2025}
}

@techreport{oecd2025behind,
  title        = {Behind ESG Ratings: Unpacking Sustainability Metrics},
  author       = {{Organisation for Economic Co-operation and Development (OECD)}},
  year         = {2025},
  institution  = {OECD Publishing},
  doi          = {10.1787/3f055f0c-en},
  url          = {https://www.oecd.org/content/dam/oecd/en/publications/reports/2025/02/behind-esg-ratings_4591b8bb/3f055f0c-en.pdf}
}

@inproceedings{zajac2023ground,
  title={Ground Truth Or Dare: Factors Affecting The Creation Of Medical Datasets For Training AI},
  author={Zajac, Hubert Dariusz and Avlona, Rozalia Natalia and Andersen, Tariq Osman and Kensing, Finn and Shklovski, Irina},
  booktitle={Proceedings of the 2023 AAAI/ACM Conference on AI, Ethics, and Society (AIES '23)},
  pages={351--362},
  year={2023},
  organization={Association for Computing Machinery},
  doi={10.1145/3600211.3604766}
}

@article{shneiderman2020human,
  title={Human-centered artificial intelligence: Reliable, safe \& trustworthy},
  author={Shneiderman, Ben},
  journal={International Journal of Human--Computer Interaction},
  volume={36},
  number={6},
  pages={495--504},
  year={2020}
}

@inproceedings{bansal2021does,
  title={Does the whole exceed its parts? The effect of AI explanations on complementary team performance},
  author={Bansal, Gagan and Wu, Tongshuang and Zhou, Joyce and Fok, Raymond and Nushi, Besmira and Kamar, Ece and Ribeiro, Marco Tulio and Weld, Daniel S.},
  booktitle={Proceedings of the 2021 CHI Conference on Human Factors in Computing Systems},
  pages={1--16},
  year={2021},
  organization={ACM},
  doi={10.1145/3411764.3445717}
}

@inproceedings{suresh2021framework,
  title={A framework for understanding sources of harm throughout the machine learning life cycle},
  author={Suresh, Harini and Guttag, John V.},
  booktitle={Proceedings of the 2021 ACM Conference on Fairness, Accountability, and Transparency},
  year={2021},
  pages={1--12},
  publisher={ACM},
  doi={10.1145/3442188.3445922}
}

@misc{wong2020ratetheraters,
  title        = {Rate the Raters 2020: Investor Survey and Interview Results},
  author       = {Wong, Christina and Petroy, Erika},
  howpublished = {SustainAbility (an ERM Group company)},
  year         = {2020},
  month        = mar,
  note         = {Accessed 2025}
}

@misc{stedman2025esgframeworks,
  author       = {Stedman, Craig and Gibbons Paul, Lauren},
  title        = {10 Top ESG Reporting Frameworks Explained and Compared},
  howpublished = {TechTarget},
  year         = {2025},
  month        = nov,
  note         = {Accessed: 2025-12-27}
}

@article{amelzadeh2018why,
  title   = {Why and How Investors Use ESG Information: Evidence from a Global Survey},
  author  = {Amel-Zadeh, Amir and Serafeim, George},
  journal = {Financial Analysts Journal},
  volume  = {74},
  number  = {3},
  pages   = {87--103},
  year    = {2018},
  doi     = {10.2469/faj.v74.n3.2}
}

@article{choi2025structuring,
  title     = {Structuring the Unstructured: A Multi-Agent System for Extracting and Querying Financial KPIs and Guidance},
  author    = {Choi, Chanyeol and Lopez-Lira, Alejandro and Lee, Yongjae and Kwon, Jihoon and Kim, Minjae and Hwang, Juneha and Ha, Minsoo and Kim, Chaewoon and Ha, Jaeseon and Yun, Suyeol and Kim, Jin},
  journal   = {arXiv preprint arXiv:2505.19197},
  year      = {2025},
  note      = {FinIR 2025},
  doi       = {10.48550/arXiv.2505.19197}
}

@article{Windolph2011,
  author       = {Windolph, Sarah Elena},
  title        = {Assessing Corporate Sustainability Through Ratings: Challenges and Their Causes},
  journal      = {Journal of Environmental Sustainability},
  volume       = {1},
  number       = {1},
  pages        = {Article 5},
  year         = {2011},
  doi          = {10.14448/jes.01.0005},
  url          = {https://repository.rit.edu/jes/vol1/iss1/5},
}

@article{DeVilliers2020Trends,
  author  = {De Villiers, Charl and Sharma, Umesh and Molinari, Marco},
  title   = {Trends in sustainability reporting: The role of regulation and reporting standards},
  journal = {Accounting, Auditing \& Accountability Journal},
  volume  = {33},
  number  = {7},
  pages   = {1545--1570},
  year    = {2020},
  doi     = {10.1108/AAAJ-05-2019-3990}
}

@article{Hatter2025Transatlantic,
  author    = {Hatter, Debra Gatison and May, Caroline and Barbone, Giulia and Ngunjiri, John and Thomas, Alexia},
  title     = {The transatlantic divide in ESG disclosure requirements: Why this matters to global businesses},
  journal   = {Business Law Today},
  year      = {2025},
  month     = {March 18},
  publisher = {American Bar Association},
  url       = {https://www.americanbar.org/groups/business_law/resources/business-law-today/2025-march/divide-in-esg-disclosure-requirements/}
}

@article{Hassani2024DiscourseEmissions,
  author       = {Hassani, Bertrand Kian and Bahini, Yacoub and Mushtaq, Rizwan},
  title        = {Discourse vs Emissions: Analysis of Corporate Narratives, Symbolic Practices, and Mimicry through Large Language Models},
  journal      = {arXiv preprint},
  year         = {2024},
  eprint       = {arXiv:2410.01222},
  archivePrefix= {arXiv},
  primaryClass = {cs.CL},
}

@article{Troshani2024SustainabilityInfrastructure,
  author    = {Troshani, Indrit and Rowbottom, Nick},
  title     = {Corporate sustainability reporting and information infrastructure},
  journal   = {Accounting, Auditing \& Accountability Journal},
  year      = {2024},
  note      = {under review / forthcoming (qualitative evidence on infrastructure constraints)},
}

@article{park2024extraterritoriality,
  title={Untangling the Extraterritoriality of ESG Regulation},
  author={Park, Stephen Kim},
  journal={North Carolina Journal of International Law},
  volume={49},
  number={4},
  year={2024}
}

@article{liu2022quantitative,
  title={Quantitative ESG disclosure and divergence of ESG ratings},
  author={Liu, Min},
  journal={Frontiers in Psychology},
  volume={13},
  pages={936798},
  year={2022},
  doi={10.3389/fpsyg.2022.936798},
  url={https://doi.org/10.3389/fpsyg.2022.936798}
}

@article{shi2025esg_divergence,
  title={ESG Rating Divergence: Existence, Driving Factors, and Impact Effects},
  author={Shi, Yong and Yao, Tongsheng},
  journal={Sustainability},
  volume={17},
  number={10},
  pages={4717},
  year={2025},
  publisher={MDPI},
  doi={10.3390/su17104717},
  url={https://doi.org/10.3390/su17104717}
}

@article{bronzini2024glitter,
  title={Glitter or gold? Deriving structured insights from sustainability reports via large language models},
  author={Bronzini, Marco and Nicolini, Carlo and Lepri, Bruno and Passerini, Andrea and Staiano, Jacopo},
  journal={EPJ Data Science},
  volume={13},
  number={1},
  pages={41},
  year={2024},
  publisher={Springer Berlin Heidelberg}
}

@inproceedings{angioni2024investigating,
  title={Investigating Environmental, Social, and Governance (ESG) Discussions in News: A Knowledge Graph Analysis Empowered by AI},
  author={Angioni, Simone and Consoli, Sergio and Dess{\'\i}, Danilo and Osborne, Francesco and Salatino, A and others},
  booktitle={CEUR WORKSHOP PROCEEDINGS},
  volume={3697},
  pages={49--60},
  year={2024},
  organization={CEUR}
}

@article{foody2024ground,
  title={Ground Truth in Classification Accuracy Assessment: Myth and Reality},
  author={Foody, Giles M.},
  journal={Geomatics},
  volume={4},
  number={1},
  pages={81--90},
  year={2024},
  month={February},
  doi={10.3390/geomatics4010005}
}

@inproceedings{snow2008cheap,
  title={Cheap and fast—but is it good? Evaluating non-expert annotations for natural language tasks},
  author={ Snow, Rion and O’Connor, Brendan and Jurafsky, Daniel and Ng, Andrew},
  booktitle={Proceedings of the 2008 Conference on Empirical Methods in Natural Language Processing},
  year={2008},
  url={https://aclanthology.org/D08-1027/}
}

@article{zhang2020automatic,
  title={Automatic information extraction from unstructured documents: A survey},
  author={ Zhong, Lingfeng and Wu, Jia and Li, Qian and Peng, Hao and Wu, Xindong},
  journal={ACM Computing Surveys (CSUR)},
  volume={53},
  number={4},
  pages={1--37},
  year={2020},
  doi={10.1145/3398034}
}
\bibliographystyle{icml2026}

\newpage
\appendix
\crefalias{section}{appendix}
\onecolumn

\section{Limitations}
The primary limitation of this research lies in its emphasis on establishing a methodological framework rather than delivering a fully implemented, end-to-end practical solution with validation. 

\begin{enumerate}

    \item \textbf{Scope of Empirical Validation}: This work focuses on establishing a methodological foundation rather than presenting a fully implemented, end-to-end system with large-scale empirical validation. Future work can focus on deploying an end-to-end STRIDE–SR-Delta system and performing validation across diverse rating providers, industries, and regulatory contexts.
    \item \textbf{Discrete Parameterization and Formula Robustness}: A few formula elements are modeled as binary indicators in the current framework. While this simplifies presentation, it limits expressiveness. A more principled approach would represent these elements on a continuous scale (e.g., entropy or information-theoretic measures) to capture graded levels of informational content and uncertainty. In addition, some definitions remain vague. For example, under the criterion “AI-driven and human-governed,” human intervention can be interpreted either as a safeguard that strengthens system reliability or as an indicator of reduced automation. This ambiguity limits the evaluative value of the criterion.
    \item \textbf{Weight Assignment}: This study does not formalize a methodology for estimating element-level weights within the STRIDE framework. Developing weighting schemes remains an important direction for future research. 
    \item \textbf{Thresholds}: In the current implementation, any strictly positive value of the STRIDE formula is treated as evidence of information in the dataset. However, no threshold function is defined to differentiate varying degrees of informational content or quality. Developing principled thresholding rules remains an important direction for future work, with implications for data set validity and downstream evaluation robustness.
    
\end{enumerate}
\newpage
\section{Equation}
\label{app:equations}
\begin{tcolorbox}[
  title={Human--Machine Trust Formulation},
  colback=white,
  colframe=black,
  fonttitle=\bfseries,
  boxrule=0.6pt,      
  arc=1pt,            
  left=4pt,
  right=4pt,
  top=2pt,
  bottom=2pt,
  before skip=4pt,    
  after skip=4pt
]

\begin{equation}
\label{ap_eq:trust2}
\begin{aligned}
\tau(x)
&=
\sigma\!\left(
\alpha_C C(x)
+
\alpha_R R(x)
+
\alpha_I I(x)
-
\alpha_S S(x)
\right),&
\sigma(z)=
\frac{1}{1+\exp(-z)} .
\end{aligned}
\end{equation}

\begin{equation}
\label{ap_eq:credibility}
C(x)
=
w_{\mathrm{IM}}^{C}\,\mathrm{IM}(x)
+
w_{\mathrm{AT}}^{C}\,\mathrm{AT}(x)
+
w_{\mathrm{ER}}^{C}\,\mathrm{ER}(x)
+
w_{\mathrm{TR}}^{C}\,\mathrm{TR}(x).
\end{equation}

\begin{equation}
\label{ap_eq:reliability}
R(x)
=
w_{\mathrm{GT}}^{R}\,\mathrm{GT}(x)
+
w_{\mathrm{SM}}^{R}\,\mathrm{SM}(x)
+
w_{\mathrm{AG}}^{R}\,\mathrm{AG}(x)
+
w_{\mathrm{SS}}^{R}\,\mathrm{SS}(x).
\end{equation}

\begin{equation}
\label{ap_eq:intimacy}
I(x)
=
w_{\mathrm{DE}}^{I}\,\mathrm{DE}(x)
+
w_{\mathrm{IF}}^{I}\,\mathrm{IF}(x)
+
w_{\mathrm{HG}}^{I}\,\mathrm{HG}(x).
\end{equation}

\begin{equation}
\label{ap_eq:selfserved}
S(x)
=
w_{\mathrm{T}}^{S}\,\mathrm{T}(x)
+
w_{\mathrm{RS}}^{S}\,\mathrm{RS}(x).
\end{equation}

\vspace{0.5em}
\noindent\textbf{Parameters.}
$\tau(x)\in(0,1)$ denotes the overall human--machine trust score for query instance $x$.
$C(x)$, $R(x)$, $I(x)$, and $S(x)$ represent perceived credibility, reliability,
human--AI intimacy, and self-served purpose, respectively, each normalized to $[0,1]$.
Coefficients $\alpha_C,\alpha_R,\alpha_I,\alpha_S \ge 0$ control the relative contribution
of each perception factor.
Weights $w_{\bullet}^{(\cdot)} \ge 0$ define convex combinations of sub-metrics and
sum to one within each component.

\vspace{0.5em}
\textbf{Sub-metrics.}
$\mathrm{IM}$ Inclusiveness and Materiality,
$\mathrm{AT}$ Auditable and Traceable Data,
$\mathrm{ER}$ Exemplary Reference,
$\mathrm{TR}$ Time Relevance;
$\mathrm{GT}$ Ground-Truth Annotation,
$\mathrm{SM}$ Statistical Methodology,
$\mathrm{AG}$ Agility with right trade-offs,
$\mathrm{SS}$ Security and Safety;
$\mathrm{DE}$ Domain-Expert-in-the-Loop,
$\mathrm{IF}$ Iterative Feedback Loop,
$\mathrm{HG}$ Human-Governed AI,
$\mathrm{T}$ Transparency,
$\mathrm{RS}$ Role Separation and Independent Oversight.

\end{tcolorbox}

\begin{tcolorbox}[
  title={Credibility Formulation},
  colback=white,
  colframe=black,
  fonttitle=\bfseries,
  boxrule=0.8pt,
  arc=2pt,
  left=6pt,
  right=6pt,
  top=6pt,
  bottom=6pt
]

\begin{equation}
\label{ap_eq:credibility_overall}
\begin{aligned}
C(x)
&=
w_{\mathrm{IM}}^{C}\,\mathrm{IM}(x)
+
w_{\mathrm{AT}}^{C}\,\mathrm{AT}(x)
+
w_{\mathrm{ER}}^{C}\,\mathrm{ER}(x)
+
w_{\mathrm{TR}}^{C}\,\mathrm{TR}(x).
\end{aligned}
\end{equation}

\vspace{0.3em}
\noindent\textbf{where}
$C(x)\in[0,1]$ is the perceived credibility score for $x$.
Weights satisfy $w_{\bullet}^{C}\ge 0$ and
$
w_{\mathrm{IM}}^{C}
+
w_{\mathrm{AT}}^{C}
+
w_{\mathrm{ER}}^{C}
+
w_{\mathrm{TR}}^{C}
=1.
$
\begin{equation}
\begin{aligned}
\mathrm{IM}(x)
=
\left[
\left(
\frac{\mathcal{C}_{\mathrm{ctry}}(x)}{N_{\mathrm{ctry}}}
\right)
\cdot
\left(
\frac{\mathcal{I}_{\mathrm{ind}}(x)}{N_{\mathrm{ind}}}
\right)
\cdot
\left(
\frac{\mathcal{S}(x)}{N_{\mathrm{std}}}
\right)
\cdot
\left(
\frac{1}{1 + \exp(x - 0.5)}
\right)
\right]^{\tfrac{1}{4}} .
\end{aligned}
\end{equation}

\noindent\textbf{where}
$\mathcal{C}_{\mathrm{ctry}}(x)$, $\mathcal{I}_{\mathrm{ind}}(x)$, and $\mathcal{S}(x)$ denote the sets of countries, industries, and standard layers covered by the evidence used to answer $x$. $N_{\mathrm{ctry}}$, $N_{\mathrm{ind}}$, and $N_{\mathrm{std}}$ represent the total numbers of countries, industries, and standard layers in scope. The variable $x \in \{0,1\}$ is a binary indicator specifying if external data are incorporated ($x=1$) or not ($x=0$).

\begin{equation}
\begin{aligned}
\mathrm{AT}(x)
&=
\frac{1}{|\mathcal{E}(x)|}
\sum_{e\in\mathcal{E}(x)}
\frac{\mathrm{aud}(e)+\mathrm{tr}(e)}{2}.
\end{aligned}
\label{eq:cred_at}
\end{equation}

\noindent\textbf{where} $\mathcal{E}(x)$ is the set of evidence items (e.g., report passages,
tables, news snippets, knowledge-graph facts) actually used to generate $\tau(x)$.

\begin{equation}
\label{ap_eq:cred_er}
\begin{aligned}
\mathrm{ER}(x)
&=
1-\left(1-\sigma\right)^{n_{\mathrm{rec}}(x)},
\qquad
\sigma\in(0,1).
\end{aligned}
\end{equation}

\vspace{0.3em}
\noindent\textbf{where}
$\sigma$ is a predefined confidence level for a single recognition event,
and $n_{\mathrm{rec}}(x)$ denotes the number of recognized endorsements or certifications
of the data source (or reporting organization) per year relevant to $x$.

\begin{equation}
\label{ap_eq:cred_tr}
\begin{aligned}
\mathrm{TR}(x)
&=
\exp\!\bigl(-\lambda\,\Delta t(x)\bigr),
\qquad
\lambda>0.
\end{aligned}
\end{equation}

\vspace{0.3em}
\noindent\textbf{where}
$\Delta t(x)$ denotes the time lag (e.g., in years) between the analysis time
and the timestamp of the key input data used to answer $x$,
and $\lambda$ controls the recency decay rate.

\end{tcolorbox}
\begin{tcolorbox}[
  title={Reliability Formulation},
  colback=white,
  colframe=black,
  fonttitle=\bfseries,
  boxrule=0.6pt,      
  arc=1pt,            
  left=4pt,
  right=4pt,
  top=2pt,
  bottom=2pt,
  before skip=4pt,    
  after skip=4pt
]

\begin{equation}
\label{ap_eq:reliability_overall2}
\begin{aligned}
R(x)
&=
w_{\mathrm{GT}}^{R}\,\mathrm{GT}(x)
+
w_{\mathrm{SM}}^{R}\,\mathrm{SM}(x)
+
w_{\mathrm{AG}}^{R}\,\mathrm{AG}(x)
+
w_{\mathrm{SS}}^{R}\,\mathrm{SS}(x).
\end{aligned}
\end{equation}

\vspace{0.3em}
\noindent\textbf{where}
$R(x)\in[0,1]$ denotes the perceived reliability score for query instance $x$.
$\mathrm{GT}$, $\mathrm{SM}$, $\mathrm{AG}$, and $\mathrm{SS}$ correspond to
Ground-Truth Annotation,
Rigorous Statistical Methodology,
Agility with right trade-offs,
and Security and Safety, respectively.
Weights satisfy $w_{\bullet}^{R}\ge 0$ and
$
w_{\mathrm{GT}}^{R}
+
w_{\mathrm{SM}}^{R}
+
w_{\mathrm{AG}}^{R}
+
w_{\mathrm{SS}}^{R}
=1.
$

\begin{equation}
\label{ap_eq:rel_sm}
\begin{aligned}
\mathrm{SM}(x)
&=
w_{\mathrm{sat}}\,
\min\!\left(1,\frac{n(x)}{n_{\mathrm{sat}}}\right)
+
w_{\mathrm{rep}}\,
\Bigl(1-\sigma_p(x)\Bigr),
\end{aligned}
\end{equation}

\noindent where
\begin{equation}
\sigma_p(x)
=
\sqrt{\frac{1}{K}\sum_{k=1}^{K}\left(\hat{p}_{x,k}-\overline{\hat{p}}_x\right)^2},
\qquad
\overline{\hat{p}}_x=\frac{1}{K}\sum_{k=1}^{K}\hat{p}_{x,k}.
\end{equation}

\noindent\textbf{where}
$n(x)$ denotes the effective sample size relevant to $x$,
$n_{\mathrm{sat}}$ is a saturation threshold beyond which marginal gains from
additional samples are negligible,
$\hat{p}_x$ represents the empirical stratified sample distribution (e.g., by
country, industry, or topic),
$p^\star$ denotes the target population distribution,
and $\mathrm{\sigma}(\hat{p}_x, p^\star)$ measures the normalized deviation between
the sample and population distributions.
The weights satisfy $w_{\mathrm{sat}}, w_{\mathrm{rep}} \ge 0$ and
$w_{\mathrm{sat}} + w_{\mathrm{rep}} = 1$.

\begin{equation}
\label{ap_eq:rel_gt}
\begin{aligned}
\mathrm{GT}(x)
&=
w_H,p_{\mathrm{human}}(x)
+
w_T,\min!\left(1,\frac{\overline{u}(x)}{u_{\max}}\right)
+
w_{HM},\sigma_{HM}(x)
+
w_{MM},\sigma_{MM}(x),
\end{aligned}
\end{equation}

\noindent \textbf{where} $p_{\mathrm{human}}(x)$ denotes the proportion of expert human annotations, $\overline{u}(x)$ is the average expert tenure (in years) with normalization cap $u_{\max}$, $\sigma_{HM}(x)$ represents the standard deviation of human--machine agreement, and $\sigma_{MM}(x)$ denotes the aggregated (summed) deviation across machine--machine agreement scores. Weights satisfy $w_H, w_T, w_{HM}, w_{MM} \ge 0$ and $w_H + w_T + w_{HM} + w_{MM} = 1$.

\begin{equation}
\label{ap_eq:rel_ag2}
\begin{aligned}
\mathrm{AG}(x)
&=
\min\!\left(
1,\;
\max\!\left(
0,\;
\frac{\Delta \mathrm{Acc}(x)}{\Delta p(x)+\epsilon}
\right)
\right),
\qquad \epsilon>0.
\end{aligned}
\end{equation}

\vspace{0.3em}
\noindent\textbf{where}
$\Delta \mathrm{Acc}(x)$ denotes the accuracy improvement after a benchmark or method update,
$\Delta p(x)\in[0,1]$ is the normalized proportion of that update,
and $\epsilon$ avoids division by zero.

\begin{equation}
\label{eq:ss_safe}
\mathrm{SS}(x)
=
1 - \frac{N_{\mathrm{harm}}(x)}{N_{\mathrm{total}}(x)} .
\end{equation}

\vspace{0.3em}
\noindent\textbf{where} $N_{\mathrm{harm}}(x)$ denotes the number of extracted rows associated with
harmful or sensitive information relevant to $x$, and $N_{\mathrm{total}}(x)$
denotes the total number of extracted rows.

\end{tcolorbox}

\begin{tcolorbox}[
  title={Intimacy (Human--AI Collaboration) Formulation},
  colback=white,
  colframe=black,
  fonttitle=\bfseries,
  boxrule=0.6pt,      
  arc=1pt,            
  left=4pt,
  right=4pt,
  top=2pt,
  bottom=2pt,
  before skip=4pt,    
  after skip=4pt
]

\begin{equation}
\label{ap_eq:intimacy_overall3}
\begin{aligned}
I(x)
&=
w_{\mathrm{HG}}^{I}\,\mathrm{HG}(x)
+
w_{\mathrm{DE}}^{I}\,\mathrm{DE}(x)
+
w_{\mathrm{IF}}^{I}\,\mathrm{IF}(x).
\end{aligned}
\end{equation}

\vspace{0.3em}
\noindent\textbf{where}
$I(x)\in[0,1]$ denotes the perceived intimacy score for query instance $x$.
Weights satisfy $w_{\bullet}^{I}\ge 0$ and
$
w_{\mathrm{DE}}^{I}
+
w_{\mathrm{IF}}^{I}
+
w_{\mathrm{HG}}^{I}
=1.
$

\begin{equation}
\label{ap_eq:human_governed3}
\begin{aligned}
\mathrm{HG}(x)
&=
\frac{n_{\mathrm{intervene}}(x)}{n_{\mathrm{cases}}(x)}.
\end{aligned}
\end{equation}

\vspace{0.3em}
\noindent\textbf{where}
$n_{\mathrm{intervene}}(x)$ denotes the number of cases in which humans intervene
(e.g., override, correct, or approve with edits),
and $n_{\mathrm{cases}}(x)$ denotes the total number of governed cases
in the evaluation window relevant to $x$.

\begin{equation}
\label{ap_eq:domain_expert2}
\mathrm{DE}(x)
=
\frac{1}{2}
\left(
\frac{\mathcal{E}(x)}{N_{\mathcal{E}}}
+
\frac{\mathcal{S}(x)}{N_{\mathcal{S}}}
\right).
\end{equation}

\vspace{0.3em}
\noindent\textbf{where}
$\mathcal{E}(x)$ denotes the set of domain experts involved in answering or validating $x$,
$\mathcal{S}(x)$ denotes the set of AI pipeline stages in which domain experts participate
(e.g., data curation, annotation, validation, evaluation, and governance),
and $N_{\mathcal{E}}$ and $N_{\mathcal{S}}$ denote the total numbers of experts and pipeline stages
in scope, respectively.

\begin{equation}
\label{ap_eq:iterative_feedback}
\mathrm{IF}(x)
=
\frac{1}{T}
\sum_{t=1}^{T}
\Delta \mathrm{Acc}_t(x),
\end{equation}

\vspace{0.3em}
\noindent\textbf{where}
$T$ denotes the total number of feedback iterations, and
$\Delta \mathrm{Acc}_t(x)$ denotes the accuracy-rate enhancement at iteration $t$
resulting from human and/or automated feedback (e.g., $\Delta \mathrm{Acc}_t(x)
=
\mathrm{Acc}_t(x)-\mathrm{Acc}_{t-1}(x)$).
\end{tcolorbox}

\begin{tcolorbox}[
  title={Self-Served Purpose Formulation},
  colback=white,
  colframe=black,
  fonttitle=\bfseries,
  boxrule=0.6pt,      
  arc=1pt,            
  left=4pt,
  right=4pt,
  top=2pt,
  bottom=2pt,
  before skip=4pt,    
  after skip=4pt
]

\begin{equation}
\label{ap_eq:selfserved_overall}
\begin{aligned}
S(x)
&=
w_{\mathrm{TL}}^{S}\,\mathrm{TL}(x)
+
w_{\mathrm{RS}}^{S}\,\mathrm{RS}(x).
\end{aligned}
\end{equation}

\vspace{0.3em}
\noindent\textbf{where}
$S(x)\in[0,1]$ denotes the perceived self-served purpose score
(lower values indicate stronger transparency and governance).
Weights satisfy $w_{\bullet}^{S}\ge 0$ and
$
w_{\mathrm{TL}}^{S}
+
w_{\mathrm{RS}}^{S}
=1.
$


\begin{equation}
T(x)
\;=\;
\frac{\left|\mathcal{A}_{\mathrm{dis}}(x)\right|}{\left|\mathcal{A}_{\mathrm{req}}(x)\right|}
\;\in\; [0,1].
\label{eq:assumption_coverage}
\end{equation}

\noindent\textbf{where} $\mathcal{A}_{\mathrm{req}}(x)$ denotes the set of required assumptions for building the benchmark dataset unit $x$, and $\mathcal{A}_{\mathrm{dis}}(x) \subseteq \mathcal{A}_{\mathrm{req}}(x)$ denotes the set of explicitly disclosed assumptions.

\begin{equation}
\label{eq:role_separation}
RS(x)
\;=\;
\frac{\left|\mathcal{E}_{\mathrm{sep}}(x)\right|}{\left|\mathcal{E}^{\star}\right|}
\;\in\; [0,1].
\end{equation}

\noindent\textbf{where} $\mathcal{R}=\{H,M,HM\}$, and let $\mathcal{E}^{\star}$ denote the set of all potential relations among roles. Let $\mathcal{E}_{\mathrm{sep}}(x) \subseteq \mathcal{E}^{\star}$ be the subset of relations whose responsibilities and handoffs are explicitly separated in unit $x$.
\end{tcolorbox}

\newpage
\section{A Case Study - Luxshare Precision Industry Co. Ltd ("Luxshare")}
\label{app:casestudy}
The STRIDE model generates two outputs: structured datasets and a quantitative readiness score for each dataset, computed using the defined formula. In this section, we demonstrate the general steps for constructing datasets using STRIDE and present an illustrative example of a dataset built under the STRIDE framework. We outline how SR-Delta can be applied to surface actionable insights. We selected Luxshare from the Fortune 500 list. It received a "BB" rating from MSCI from November 14th, 2023, to May 29th, 2024. We use Luxshare's 2024 sustainability report as primary input, as it covers the majority of the analysis period. The final STRIDE score for the Luxshare dataset is 0.56. In our analysis using SR-Delta, ambiguities in disclosure definitions and over- or under-penalization could lead to rating adjustments. These adjustments include a potential upward adjustment of +1.2 for the former, a downward adjustment of –1.1 to -0.5 for the latter, with a net adjustment between +0.1 to +0.7. 

\subsection{Credibility - Raw Input Data Selection}

\begin{figure}[ht]
    \centering
    \includegraphics[width=1\linewidth]{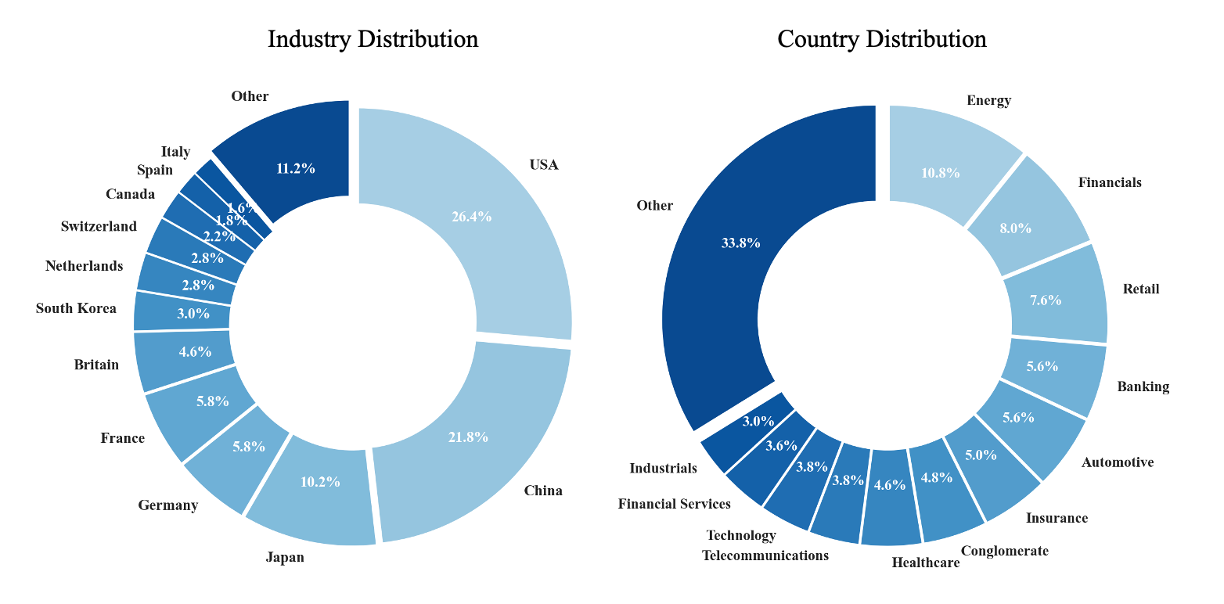}
    \caption{The Fortune Global 500 comprises firms with diversity across geographic regions and industry sectors.}
    \label{fig:company_distribution}
\end{figure}

\textbf{Inclusiveness and Materiality}
For values, it is important to capture signals across sectors and regions. Companies from the Fortune 500, as listed on kaggle.com for the year 2025 \footnote{See \url{https://www.kaggle.com/datasets/edgarhuichen/fortune-global-500}}, can serve as proxies. We gathered links to online sustainability reports from Fortune Global 500 companies for 2020-2024. The data set includes 2,000 reports covering 75 industries and companies headquartered in 35 countries. Although the sample is not evenly distributed across industries or regions, it captures substantial heterogeneity and offers coverage of representative firms across sectors and geographies (\Cref{fig:company_distribution}). 

\begin{figure}[H]
    \centering
    \includegraphics[width=1\linewidth]{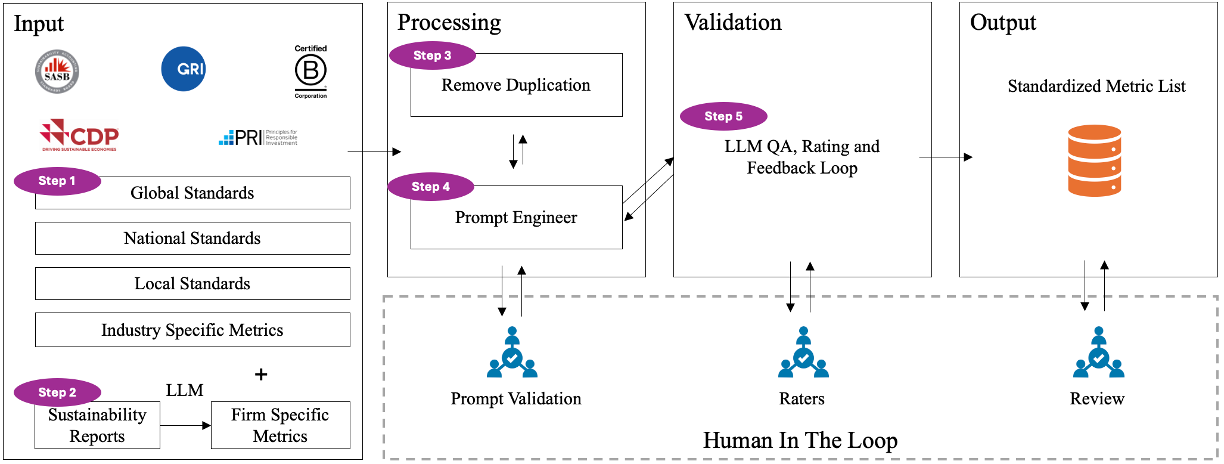}
    \caption{In our case study, metrics across all levels are consolidated and de-duplicated using LLMs. We recommend incorporating humans in the loop as suggested in the dashed box.}
    \label{fig:metrics_generator}
\end{figure}

When constructing the metrics list, it is essential to include international, national, local, industry-specific, and firm-specific metrics. Luxshare, as an example, operates in at least 12 countries and is in the electronics manufacturing industry, according to Luxshare Precision's sustainability reports. We use the standards and principles listed in \Cref{tab:firstfour_metrics} as inputs for the first four metric levels. As illustrated in \Cref{fig:metrics_generator}, we first use LLMs to integrate a predefined metrics list for the first four categories (Step 1). For company-specific metrics, we employ LLMs to generate a custom metric list in an unsupervised manner (Step 2). The two lists are then merged and de-duplicated (Step 3) using prompts (Step 4), followed by LLM-based verification to produce the final metric set (Step 5). This process yields a consolidated list of 527 metrics. We recommend incorporating human-in-the-loop safeguards for metric generation, prompt-based metric evaluation, and metric review, as indicated by the grey dashed box, particularly when additional oversight or risk mitigation is required. For the purposes of the case study, we do not use external data as metric inputs or value inputs. 

\begin{table}[htbp]

\caption{Sustainability Reporting and Disclosure Frameworks Across Four Levels}
\label{tab:firstfour_metrics}
\centering
\begin{tabularx}{\textwidth}{p{2cm} p{4.5cm} X}
\toprule
\textbf{Level} & \textbf{Description} & \textbf{Frameworks and Standards} \\
\midrule
\textbf{Global} 
& International standards
& Global Reporting Initiative (GRI); 
Task Force on Climate-related Financial Disclosures (TCFD); 
IFRS Sustainability Disclosure Standards (ISSB); 
UN Sustainable Development Goals (UN SDGs); \\

\midrule
\textbf{National} 
& Country-specific regulations 
& UK Sustainability Reporting Standards; K-ESG Guidelines (Korea); 
SGX Core ESG Metrics (Singapore); 
Corporate Sustainability Disclosure Standards (China) \\

\midrule
\textbf{Local} 
& Sub-national standards
& Implementation Guidance on Climate Information Disclosure
under the Hong Kong Exchange Environmental, Social and Governance (ESG) Framework;  Beijing Ecological Environment Status Report
\\

\midrule
\textbf{Industry} 
& Sector-specific standards
& Criteria for the Sustainability Assessment of
Network Equipment for the Global Electronics\\

\bottomrule
\end{tabularx}

\end{table}

\begin{tcolorbox}[
  title={Integrity Measure (IM) Luxshare Score},
  colback=white,
  colframe=gray!70,
  fonttitle=\bfseries,
  boxrule=0.6pt,
  arc=1pt,
  left=4pt,
  right=4pt,
  top=2pt,
  bottom=2pt,
  before skip=4pt,
  after skip=4pt
]

\begin{equation}
\label{eq:im_luxshare}
\mathrm{IM}(x)
=
\left[
\left(
\frac{\mathcal{C}_{\mathrm{ctry}}(x)}{N_{\mathrm{ctry}}}
\right)
\cdot
\left(
\frac{\mathcal{I}_{\mathrm{ind}}(x)}{N_{\mathrm{ind}}}
\right)
\cdot
\left(
\frac{\mathcal{S}(x)}{N_{\mathrm{std}}}
\right)
\cdot
\left(
\frac{1}{1 + \exp(x - 0.5)}
\right)
\right]^{\tfrac{1}{4}} .
\end{equation}

\vspace{0.3em}
\noindent\textbf{Illustrative Inputs (per Eq.~(12)):}
\begin{itemize}
    \item $\mathcal{C}_{\mathrm{ctry}}(x) = 35$, number of countries covered; \quad $N_{\mathrm{ctry}} = 195$, total number of countries;
    \item $\mathcal{I}_{\mathrm{ind}}(x) = 75$, number of industries covered; \quad $N_{\mathrm{ind}} = 163$, total number of industries;
    \item $\mathcal{S}(x) = 5$, number of standard layers included; \quad $N_{\mathrm{std}} = 5$, total number of standard layers;
    \item $x = 0$, indicating that no external data are incorporated.
\end{itemize}

\vspace{0.3em}
\begin{equation*}
\label{eq:im_zero}
\mathrm{IM}(x)
=
\left[
\left(
\frac{35}{195}
\right)
\cdot
\left(
\frac{75}{163}
\right)
\cdot
\left(
\frac{5}{5}
\right)
\cdot
\left(
\frac{1}{1 + \exp(0 - 0.5)}
\right)
\right]^{\tfrac{1}{4}}
\;\approx\; 0.48.
\end{equation*}
\end{tcolorbox}

\textbf{Auditable and Traceable Data} The data and model process defined for Luxshare constitutes a limited-scope case study. Full traceability and auditability are not directly applicable in this setting because data inputs are manually curated, the analytical steps are explicitly defined, and the pipeline is linear rather than distributed. There is no complex provenance chain or system-level uncertainty that would require formal traceability or independent audit mechanisms.

\textbf{Exemplary Reference} During the 2024 fiscal year, Luxshare received 18 awards in its sustainability report. Assume that the confidence of earning the reward is 80\% since they have been able to receive the same awards over years, 

\begin{tcolorbox}
[
  title={Exemplary Reference (ER) Score},
  colback=white,
  colframe=gray!70,
  fonttitle=\bfseries,
  boxrule=0.6pt,
  arc=1pt,
  left=4pt,
  right=4pt,
  top=2pt,
  bottom=2pt,
  before skip=4pt,
  after skip=4pt
]
\begin{equation*}
\label{eq:er_def}
\mathrm{ER}(x)
=
1-\left(1-\sigma\right)^{n_{\mathrm{rec}}(x)},
\qquad
\sigma\in(0,1).
\end{equation*}

\vspace{0.3em}
\noindent\textbf{Illustrative Inputs:}
\begin{itemize}
    \item $n_{\mathrm{rec}}(x)=18$, number of recognized awards;
    \item $\sigma=0.80$, confidence level (assumed).
\end{itemize}

\vspace{0.3em}
\begin{equation*}
\label{eq:er_example}
\mathrm{ER}(x)
=
1-\left(1-0.80\right)^{18}
=
1-(0.20)^{18}
\;\approx\; 1.00.
\end{equation*}
\end{tcolorbox}

\vspace{0.6em}
\textbf{Time Relevance}
The MSCI case-study report covers Novermber 14th 2023 to May 29th 2024, while the sustainability reports overs January 1st 2024 to December 31st, 2024, resulting in an overlap of approximately five months.

\begin{tcolorbox}
[
  title={Time Relevance (TR) Score},
  colback=white,
  colframe=gray!70,
  fonttitle=\bfseries,
  boxrule=0.6pt,
  arc=1pt,
  left=4pt,
  right=4pt,
  top=2pt,
  bottom=2pt,
  before skip=4pt,
  after skip=4pt
]

\begin{equation*}
\label{eq:tr_def}
\mathrm{TR}(x)
=
\exp\!\bigl(-\lambda\,\Delta t(x)\bigr),
\qquad
\lambda>0.
\end{equation*}

\vspace{0.3em}
\noindent\textbf{Illustrative Inputs:}
\begin{itemize}
    \item $\lambda = 0.1$, decay parameter (assumed);
    \item $\Delta t(x) = \tfrac{5}{12}$, time lag in years (five months).
\end{itemize}

\vspace{0.3em}
\begin{equation*}
\label{eq:tr_example}
\mathrm{TR}(x)
=
\exp\!\Bigl(-0.1 \cdot \tfrac{5}{12}\Bigr)
=
\exp\!\bigl(-0.4167\bigr)
\;\approx\; 0.96.
\end{equation*}

\end{tcolorbox}
\vspace{0.6em}
Given the above calculations, the total credibility score is calculated as follows.

\begin{tcolorbox}
[
  title={Total Credibility (C) Score},
  colback=white,
  colframe=black,
  fonttitle=\bfseries,
  boxrule=0.6pt,
  arc=1pt,
  left=4pt,
  right=4pt,
  top=2pt,
  bottom=2pt,
  before skip=4pt,
  after skip=4pt
]

\vspace{0.4em}
\begin{equation*}
\label{eq:cred_def}
C(x)
=
w_{\mathrm{IM}}^{C}\,\mathrm{IM}(x)
+
w_{\mathrm{AT}}^{C}\,\mathrm{AT}(x)
+
w_{\mathrm{ER}}^{C}\,\mathrm{ER}(x)
+
w_{\mathrm{TR}}^{C}\,\mathrm{TR}(x),
\end{equation*}

\vspace{0.3em}
\noindent\textbf{Illustrative Inputs (Equal Weights and AT is not applicable):}
\begin{itemize}
    \item $w_{\mathrm{IM}}^{C} = w_{\mathrm{ER}}^{C} = w_{\mathrm{TR}}^{C} = 1/3$;
    \item $\mathrm{IM}(x) = 0.48$ (Materiality and Inclusiveness);
     \item $\mathrm{AT}(x)$ is not applicable;
    \item $\mathrm{ER}(x) = 1.00$ (Evidence Recognition);
    \item $\mathrm{TR}(x) = 0.96$ (Time Relevance).
\end{itemize}

\vspace{0.3em}
\begin{equation*}
\label{eq:cred_example}
C(x) = 1/3 \cdot 0.48 + 1/3 \cdot 1.00 + 1/3 \cdot 0.96 \;\approx\; 0.81    .
\end{equation*}
\end{tcolorbox}

\subsection{Reliability - Modele Performance}

\textbf{Rigorous Statistical Methodology } In the sampling process, if full datasets would be generated, ground-truth data denote a collection of manually annotated datasets, like the Luxshare dataset. They serve as supervisory input for training the LLM to generate the remaining datasets. Therefore, from the full cohort of selected companies that span four years, a representative subset are curated to reflect the broader population. To ensure the representativeness of the ground truth dataset, the sample selection process was guided by three core criteria in \Cref{tab:criteria_definitions}. \Cref{tab:dataset_metrics} provides an overlook for all selected criterion metrics.

\begin{table}[H]
\centering
\renewcommand{\arraystretch}{1.25}
\setlength{\tabcolsep}{8pt}
\caption{Conceptual Definitions of Core Dataset Criteria}
\label{tab:criteria_definitions}

\begin{tabular}{p{3cm} p{13cm}}
\hline
\textbf{Criteria} & \textbf{Definition} \\
\hline

Data Properties &
It characterizes the structural, textual, and visual attributes of a document. These metrics serve as proxies for reporting depth, information richness, and disclosure quality. \\[0.8em]
\midrule
Company Diversity &
It represents core organizational characteristics, such as scale, workforce size, and maturity. It captures structural heterogeneity across firms that shapes reporting capacity, disclosure practices, and sustainability performance comparability. \\[0.8em]
\midrule
Domain Attributes &
It represents the extent to which a company's sustainability reporting engages with environmental, social, and governance domains. It captures variation in topical focus, narrative depth, and standards alignment that shapes disclosure quality. \\

\hline
\end{tabular}
\end{table}

\begin{table}[H]
\centering
\caption{Dataset Criteria and Metric Definitions}
\label{tab:dataset_metrics}
\renewcommand{\arraystretch}{1.25}
\begin{tabular}{p{3cm} p{3.5cm} p{9.25cm}}
\hline
\textbf{Criteria} & \textbf{Metric Name} & \textbf{Metric Definition} \\
\hline

\multirow[t]{7}{*}{\textbf{Data Properties}} 
& page\_count & Total number of pages in the sustainability report. \\
& word\_count & Total number of words in the report text. \\
& sentence\_count & Total number of sentences extracted from the report. \\
& avg\_sentence\_length & Average number of words per sentence. \\
& file\_size\_mb & File size of the report in megabytes (MB). \\
& image\_count & Total number of figures, charts, or visual elements. \\
& visual\_elements\_per\_page & Average number of visual elements (e.g., images, tables, graphs) per page. \\
\hline

\multirow[t]{6}{*}{\textbf{Company Diversity}} 
& region & Primary geographical region of the reporting entity. \\
& industry & Industry classification of the company. \\
& revenue\_billions & Annual company revenue measured in USD billions. \\
& employees\_thousands & Number of employees measured in thousands. \\
& years\_since\_founded & Number of years since the company was founded. \\
& public & Binary indicator denoting whether the company is publicly traded or privately held. \\
\hline

\multirow[t]{6}{*}{\textbf{Domain Attributes}} 
& env\_keyword\_count & Frequency of environmental-related keywords (e.g., climate, carbon, emissions) in the report text. \\
& social\_keyword\_count & Frequency of social-related keywords (e.g., diversity, labor practices, human rights) in the report text. \\
& gov\_keyword\_count & Frequency of governance-related keywords (e.g., board oversight, compliance) in the report text. \\
& framework\_count & Number of recognized sustainability reporting frameworks referenced (e.g., GRI, SASB, TCFD, SDGs). \\
& frameworks & List of sustainability reporting frameworks explicitly referenced in the report. \\
\hline

\end{tabular}
\end{table}

\begin{figure} [H]
    \centering
    \includegraphics[width=1\linewidth]{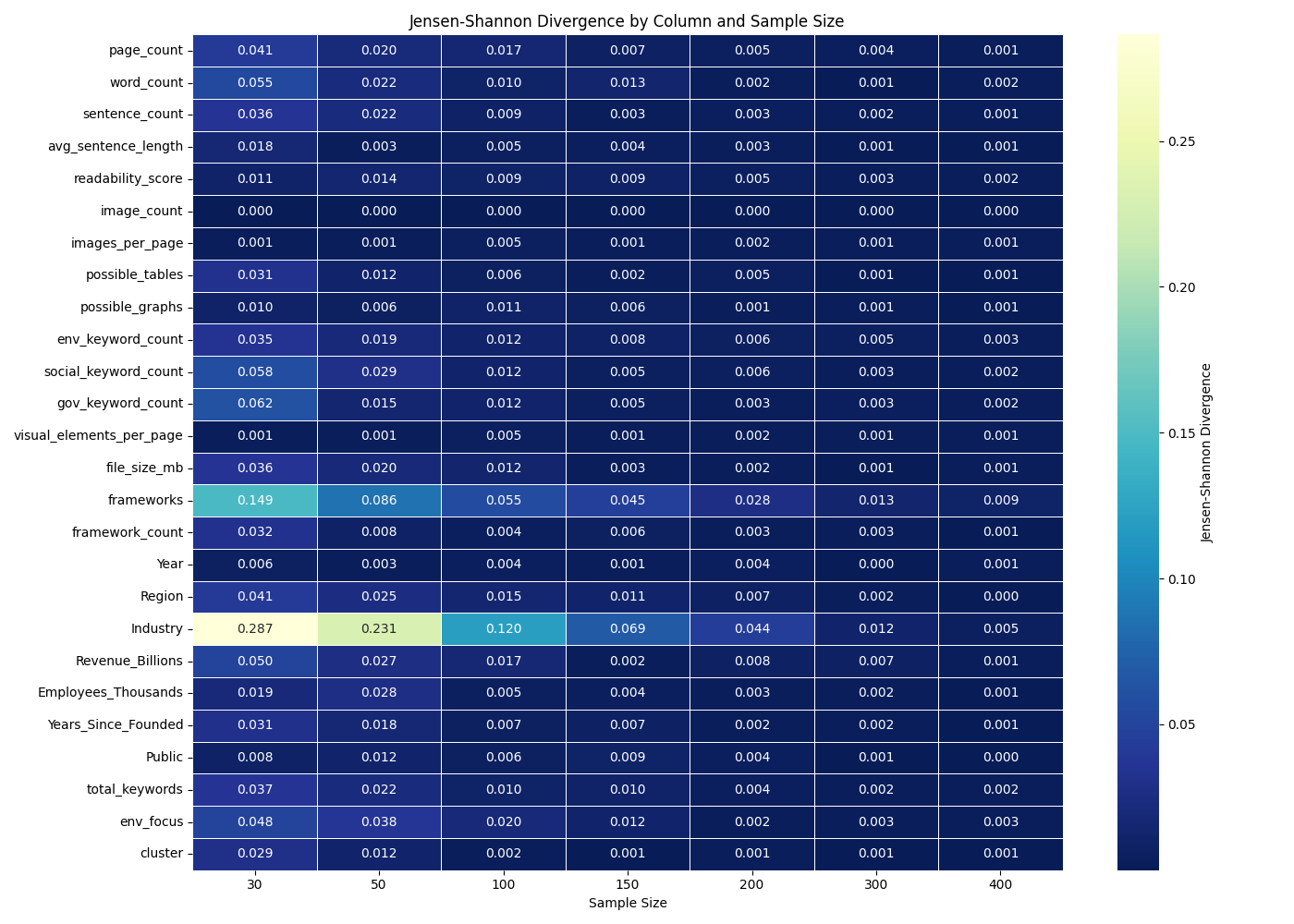}
    \caption{The largest divergence occurs in adopted sustainability frameworks and industry alignment. At a sample size of 300, their Jensen–Shannon Divergence drops to 0.013 and 0.012, indicating distributional stability and sample representativeness.}
    \label{fig:js_divergence}
\end{figure}

\begin{figure} [H]
    \centering
 
    \includegraphics[width=1\linewidth]{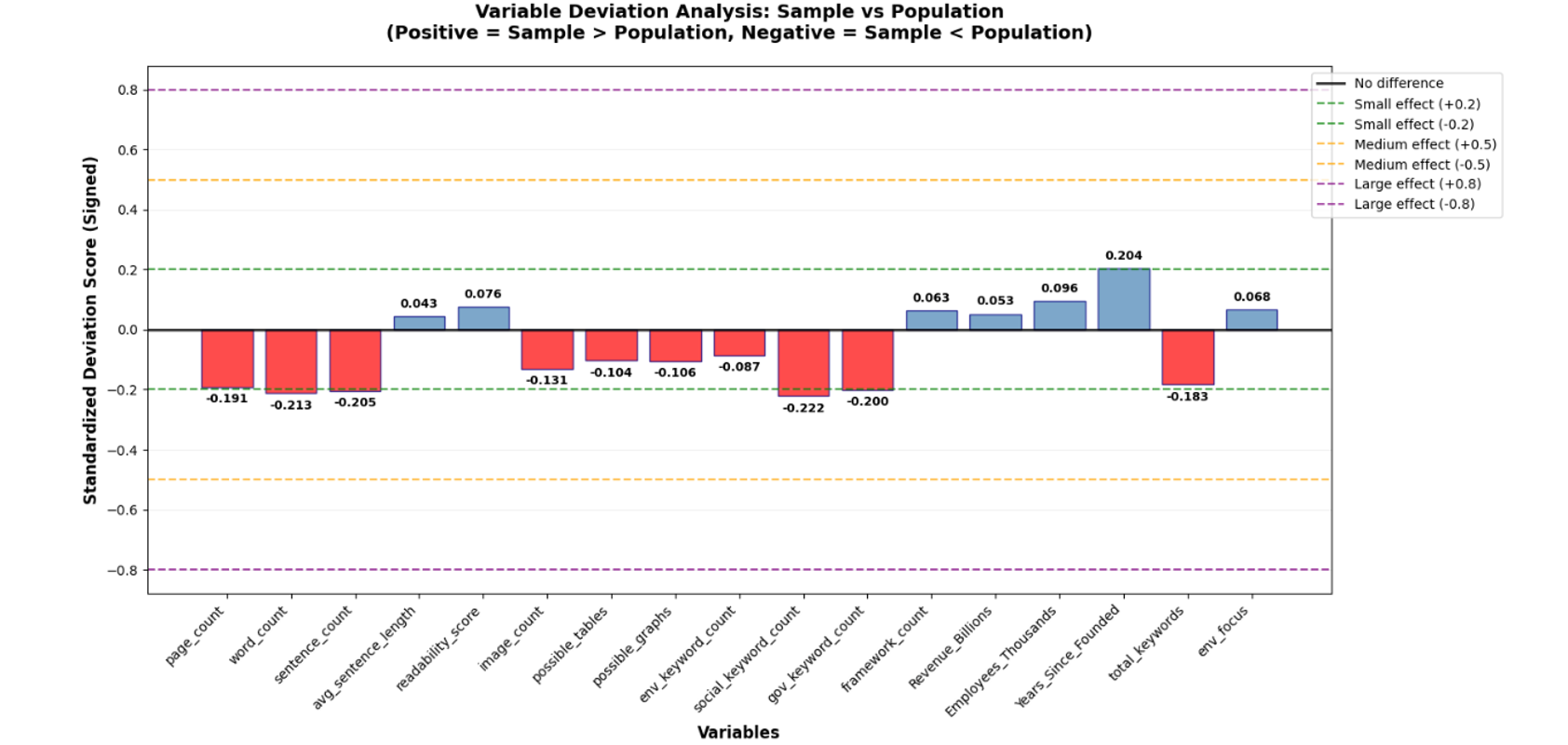} 
    \caption{The graph illustrates, by way of example, the aggregate minimum deviation from the population for the selected samples.}
    \label{fig:RS_standard_deviation}
\end{figure}

To assess sample size adequacy, we employ Jensen–Shannon divergence \Cref{fig:js_divergence} as a measure of distributional convergence. Empirically, the divergence decreases and plateaus as the sample size approaches approximately 300, indicating diminishing marginal information gain from additional samples. This stabilization suggests that a sample size of around 300 is sufficient. If all samples have been selected, the sample saturation rate is 1. We use these metrics to quantify the deviation of candidate samples from the target distribution and select the subset that minimizes the overall deviation from the population \Cref{fig:RS_standard_deviation}. As this procedure was not applicable to the Luxshare case study, the corresponding scores SM(x) is not computed.

\textbf{Ground Truth Annotation} For the Luxshare case study, metric values were initially generated and evaluated by three independent AI agents. Their outputs were aggregated by averaging to obtain a preliminary score for each metric and to compute deviation-based agreement measures. Subsequently, a domain expert with over five years of relevant experience conducted a full-dataset human annotation covering 527 metrics, providing authoritative validation and serving as the reference for downstream analysis.

\begin{tcolorbox}[
  title={Ground Truth Annotation (GT) Score},
  colback=white,
  colframe=gray!70,
  fonttitle=\bfseries,
  boxrule=0.6pt,
  arc=1pt,
  left=4pt,
  right=4pt,
  top=2pt,
  bottom=0pt,        
  before skip=4pt,
  after skip=4pt
]

\begin{equation*}
\label{ap_eq:rel_gt2}
\mathrm{GT}(x)
=
w_H\,p_{\mathrm{human}}(x)
+
w_T\,\min\!\left(1,\frac{\overline{u}(x)}{u_{\max}}\right)
+
w_{HM}\,\sigma_{HM}(x)
+
w_{MM}\,\sigma_{MM}(x).
\end{equation*}

\noindent\textbf{Illustrative Inputs:}
\begin{itemize}\setlength{\itemsep}{0.2em}\setlength{\topsep}{0pt}
    \item $w_{\mathrm{H}} = w_{\mathrm{T}} = w_{\mathrm{HM}} = w_{\mathrm{MM}} = \tfrac{1}{4}$;
    \item $p_{\mathrm{human}}(x) = \tfrac{527}{527} = 1$;
    \item $\min\!\left(1,\tfrac{\overline{u}(x)}{u_{\max}}\right) = 1$;
    \item $\sigma_{HM} = 1.43$;
    \item $\sigma_{MM} = 1.24$.
\end{itemize}

\begin{equation*}
\label{ap_eq:rel_gt_num}
\mathrm{GT}(x)
=
\frac{1}{4}\!\left(1 + 1 + 1.43 + 1.24\right)
\approx 1.17.
\end{equation*}

\end{tcolorbox}

\vspace{0.6em}
\textbf{Agility with the Right Trade-off} We deploy optimization agents that dynamically switch among multiple LLMs during the extraction and scoring process. In the Luxshare case study, the agents performed three model-switching iterations, adaptively balancing accuracy, reasoning capability, and computational efficiency. This process resulted in an average accuracy improvement of approximately 3.2\% per switch across 527 metrics. The model pool consisted of GPT-4.1, Gemini 2.5 Pro, and Grok, allowing the system to leverage complementary strengths across models and mitigate systematic biases associated with any single architecture.
\begin{tcolorbox}[
  title={Agility (AG) with the Right Trade-off Score},
  colback=white,
  colframe=gray!70,
  fonttitle=\bfseries,
  boxrule=0.6pt,
  arc=1pt,
  left=4pt,
  right=4pt,
  top=2pt,
  bottom=0pt,        
  before skip=4pt,
  after skip=4pt
]

\begin{equation*}
\label{ap_eq:rel_ag}
\mathrm{AG}(x)
=
\min\!\left(
1,\;
\max\!\left(
0,\;
\frac{\Delta \mathrm{Acc}(x)}{\Delta p(x)+\epsilon}
\right)
\right).
\end{equation*}

\noindent\textbf{Illustrative Inputs:}
\begin{itemize}\setlength{\itemsep}{0.2em}\setlength{\topsep}{0pt}
    \item $\Delta \mathrm{Acc}(x) = 3.2\% = 0.032$;
    \item $\Delta p(x) = 10\% = 0.10$;
    \item $\epsilon = 10^{-6}$.
\end{itemize}

\begin{equation*}
\label{ap_eq:rel_ag_num}
\mathrm{AG}(x)
=
\min\!\left(
1,\;
\frac{0.032}{0.10 + 10^{-6}}
\right)
\approx
\min(1,\; 0.3199968)
=
0.32.
\end{equation*}

\end{tcolorbox}

\vspace{0.6em}
\textbf{Safety and Security}
After extracting all values, we manually reviewed the populated entries and did not find harmful or insecure information.
\begin{tcolorbox}[
  title={Safety and Security Score (SS)},
  colback=white,
  colframe=gray!70,
  fonttitle=\bfseries,
  boxrule=0.6pt,
  arc=1pt,
  left=4pt,
  right=4pt,
  top=2pt,
  bottom=0pt,        
  before skip=4pt,
  after skip=4pt
]

\begin{equation*}
\label{ap_eq:rel_ss}
\mathrm{SS}(x)
=
1 - \frac{N_{\mathrm{harm}}(x)}{N_{\mathrm{total}}(x)} .
\end{equation*}

\noindent\textbf{Illustrative Inputs:}
\begin{itemize}\setlength{\itemsep}{0.2em}\setlength{\topsep}{0pt}
    \item $N_{\mathrm{harm}}(x) = 0$;
    \item $N_{\mathrm{total}}(x) = 527$.
\end{itemize}

\begin{equation*}
\label{ap_eq:rel_ss_num}
\mathrm{SS}(x)
=
1 - \frac{0}{527}
=
1.
\end{equation*}
\end{tcolorbox}

\vspace{0.6em}
Given the above calculations, the total reliability score is calculated as follows.
\begin{tcolorbox}
[
  title={Total Reliability (R) Score},
  colback=white,
  colframe=black,
  fonttitle=\bfseries,
  boxrule=0.6pt,
  arc=1pt,
  left=4pt,
  right=4pt,
  top=2pt,
  bottom=2pt,
  before skip=4pt,
  after skip=4pt
]

\vspace{0.4em}
\begin{equation*}
\label{ap_eq:reliability_overall}
\begin{aligned}
R(x)
&=
w_{\mathrm{GT}}^{R}\,\mathrm{GT}(x)
+
w_{\mathrm{SM}}^{R}\,\mathrm{SM}(x)
+
w_{\mathrm{AG}}^{R}\,\mathrm{AG}(x)
+
w_{\mathrm{SS}}^{R}\,\mathrm{SS}(x).
\end{aligned}
\end{equation*}

\vspace{0.3em}
\noindent\textbf{Illustrative Inputs (Equal Weights and SM is not applicable):}
\begin{itemize}

    \item $w_{\mathrm{GT}}^{R} = w_{\mathrm{SM}}^{R} = w_{\mathrm{SS}}^{R} = 1/3$;
    \item $\mathrm{SM}(x)$ is not applicable;
    \item $\mathrm{GT}(x) = 1.17$ (Ground Truth Annotation);
    \item $\mathrm{AG}(x) = 0.32$ (Agility);
    \item $\mathrm{SS}(x) = 1.00$ (Safety and Security Score).
\end{itemize}

\vspace{0.3em}
\begin{equation*}
\label{eq:reliability_example}
R(x)
=
\frac{1}{3}\cdot 1.17
+
\frac{1}{3}\cdot 0.32
+
\frac{1}{3}\cdot 1.00
\;\approx\; 0.83.
\end{equation*}

\end{tcolorbox}

\subsection{Intimacy}
\begin{figure*}[ht]
    \centering
    \includegraphics[width=\linewidth]{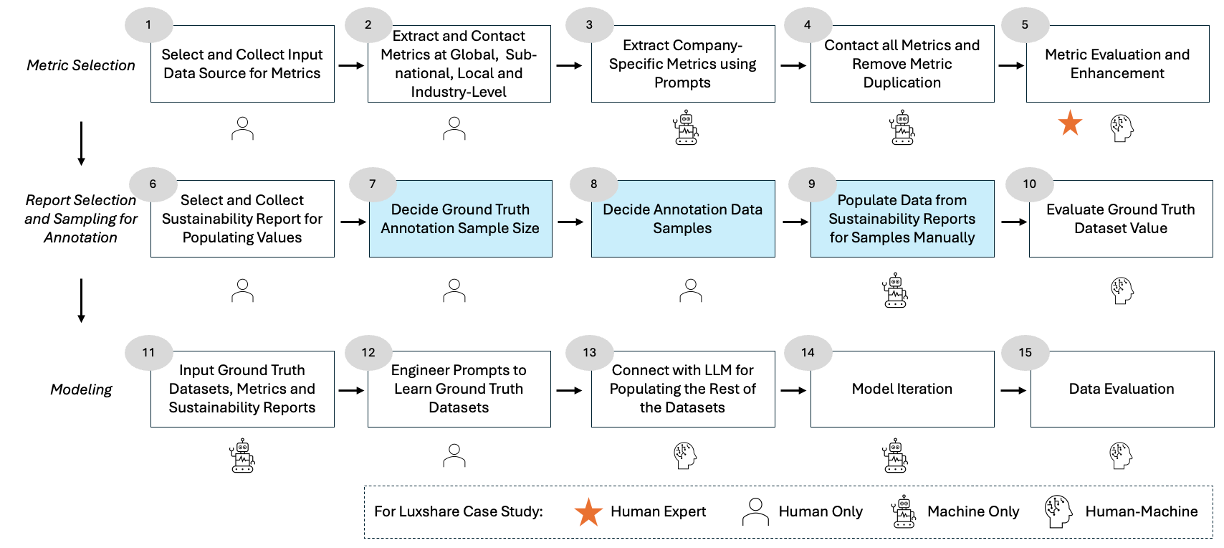}
    \caption{The figure illustrates the end-to-end workflow for metric selection, report selection and sampling for annotation, and modeling. Steps (1–5) consolidate and de-duplicate sustainability metrics across global, sub-national, local, industry, and company levels. Steps (6–10) define report selection, sampling strategy, and human ground-truth annotation. Steps (11–15) integrate annotated datasets with large language models for scalable population, iterative refinement, and data evaluation.}
    \label{fig:trust_equation}
\end{figure*}

\textbf{AI-Driven and Human-Govern} 
An AI-driven, human-governed framework is used to assess the depth of human–AI collaboration. The \Cref{fig:trust_equation} maps the Luxshare workflow and classifies each step as human-only, machine-only, or human–machine collaborative. For the Luxshare case, during dataset generation with ground-truth inputs, human interventions were introduced through prompt refinement, supplementary data injection, and rating output corrections for a total of 56 records. These interventions were concentrated in workflow stages 5, 12, and 13, which correspond to critical control points for extraction accuracy, data alignment, and scoring integrity.

\begin{tcolorbox}[
  title={Human-Governed Score (HG)},
  colback=white,
  colframe=gray!70,
  fonttitle=\bfseries,
  boxrule=0.6pt,
  arc=1pt,
  left=4pt,
  right=4pt,
  top=2pt,
  bottom=0pt,        
  before skip=4pt,
  after skip=4pt
]

\begin{equation*}
\label{ap_eq:human_governed2}
\mathrm{HG}(x)
=
\frac{n_{\mathrm{intervene}}(x)}{n_{\mathrm{cases}}(x)}.
\end{equation*}

\noindent\textbf{Illustrative Inputs:}
\begin{itemize}\setlength{\itemsep}{0.2em}\setlength{\topsep}{0pt}
    \item $n_{\mathrm{intervene}}(x) = 56$;
    \item $n_{\mathrm{cases}}(x) = 527$.
\end{itemize}

\begin{equation*}
\label{ap_eq:human_governed_num}
\mathrm{HG}(x)
=
\frac{56}{527}
\approx
0.106.
\end{equation*}

\end{tcolorbox}

\textbf{Domain Expert in the Loop} 
For the Luxshare case study, one expert-level human was involved in total for metric validation during the workflow for one process, step 5, out of the 15 steps.

\begin{tcolorbox}
[
  title={Domain Expert-in-the-Loop (DE) Score},
  colback=white,
  colframe=gray!70,
  fonttitle=\bfseries,
  boxrule=0.6pt,
  arc=1pt,
  left=4pt,
  right=4pt,
  top=2pt,
  bottom=2pt,
  before skip=4pt,
  after skip=4pt
]

\begin{equation*}
\label{ap_eq:domain_expert}
\mathrm{DE}(x)
=
\frac{1}{2}
\left(
\frac{\mathcal{E}(x)}{N_{\mathcal{E}}}
+
\frac{\mathcal{S}(x)}{N_{\mathcal{S}}}
\right).
\end{equation*}

\vspace{0.3em}
\noindent\textbf{Illustrative Inputs:}

\begin{itemize}\setlength{\itemsep}{0.2em}\setlength{\topsep}{0pt}
    \item $\mathcal{E}(x) = 1$; 
    \item $N_{\mathcal{E}} = 1$;
     \item $\mathcal{S}(x) = 1$; 
     \item $N_{\mathcal{S}} = 15$.
\end{itemize}

\vspace{0.2em}

\begin{equation*}
\label{ap_eq:domain_expert_num}
\mathrm{DE}(x)
=
\frac{1}{2}
\left(
\frac{1}{1}
+
\frac{1}{15}
\right)
\;\approx\; 0.53.
\end{equation*}
\end{tcolorbox}

\textbf{Iterative Feedback Loop} 
In the case of Luxshare, no iterative feedback loop is established, as the human annotation process is conducted as a one-off exercise and the resulting outputs are not fed back into the model. Consequently, this criterion is not applicable in the current setting.
\newpage
\begin{tcolorbox}
[
  title={Total Intimacy (I) Score},
  colback=white,
  colframe=black,
  fonttitle=\bfseries,
  boxrule=0.6pt,
  arc=1pt,
  left=4pt,
  right=4pt,
  top=2pt,
  bottom=2pt,
  before skip=4pt,
  after skip=4pt
]

\vspace{0.4em}
\begin{equation*}
\label{ap_eq:intimacy_overall}
\begin{aligned}
I(x)
&=
w_{\mathrm{DE}}^{I}\,\mathrm{DE}(x)
+
w_{\mathrm{IF}}^{I}\,\mathrm{IF}(x)
+
w_{\mathrm{HG}}^{I}\,\mathrm{HG}(x).
\end{aligned}
\end{equation*}

\vspace{0.3em}
\noindent\textbf{Illustrative Inputs (Equal Weights and IF is not applicable):}
\begin{itemize}
    \item $w_{\mathrm{DE}}^{I} = w_{\mathrm{HG}}^{I} = 1/2$;
    \item $\mathrm{IF}(x)$ is not applicable;
    \item $\mathrm{HG}(x) \approx 0.106$ (Human-Governed);
    \item $\mathrm{DE}(x) = 0.53$ (Domain Expert-in-the-Loop).
 
\end{itemize}

\vspace{0.3em}
\begin{equation*}
\label{ap_eq:intimacy_example}
I(x)
=
\frac{1}{2}\cdot 0.106
+
\frac{1}{2}\cdot 0.53
\;\approx\; 0.32.
\end{equation*}

\end{tcolorbox}

\subsection{Self-Served Purpose}

\textbf{Transparency} As the Luxshare analysis is a single-case study within the population dataset, no generalization to the broader dataset is assumed. therefore, this criterion is not applicable.

\textbf{Role Separation and Independent Oversight}

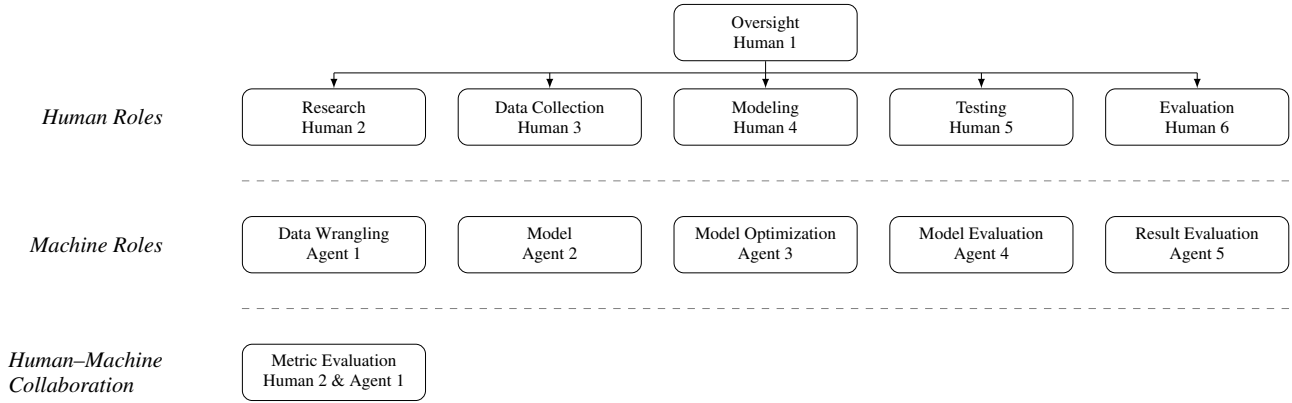
\begin{figure}[ht]
\centering
\begin{adjustbox}{max width=\linewidth}
\begin{tikzpicture}[
  box/.style={
    draw, rectangle, rounded corners,
    minimum width=2.6cm,
    minimum height=0.8cm,
    align=center,
    font=\scriptsize
  },
  labelstyle/.style={draw=none, font=\footnotesize\itshape, align=left},
  dashedline/.style={dashed, gray},
  ptr/.style={-latex, line width=0.35pt},
  line/.style={line width=0.35pt}
]

\node[box] (research) {Research\\Human 2};
\node[box, right=0.45cm of research] (datacollection) {Data Collection\\Human 3};
\node[box, right=0.45cm of datacollection] (modeling) {Modeling\\Human 4};
\node[box, right=0.45cm of modeling] (testing) {Testing\\Human 5};
\node[box, right=0.45cm of testing] (evaluation) {Evaluation\\Human 6};

\node[labelstyle, left=1.0cm of research] {Human Roles};

\node[box, above=1.2cm of modeling.west, anchor=west] (oversight) {Oversight\\Human 1};

\coordinate (HbusL) at ([yshift=0.22cm]research.north);
\coordinate (HbusR) at ([yshift=0.22cm]evaluation.north);
\coordinate (HbusMid) at ($(HbusL)!0.5!(HbusR)$);

\draw[line] (oversight.south) -- (HbusMid);
\draw[line] (HbusL) -- (HbusR);

\foreach \b in {research,datacollection,modeling,testing,evaluation}{
  \draw[ptr] ([yshift=0.22cm]\b.north) -- (\b.north);
}

\draw[dashedline] ([yshift=-0.50cm]research.south west |- research.south)
  -- ([yshift=-0.50cm]evaluation.south east |- evaluation.south);

\node[box, below=1.00cm of research] (datawrangling) {Data Wrangling\\Agent 1};
\node[box, right=0.45cm of datawrangling] (model) {Model\\Agent 2};
\node[box, right=0.45cm of model] (optimization) {Model Optimization\\Agent 3};
\node[box, right=0.45cm of optimization] (modeleval) {Model Evaluation\\Agent 4};
\node[box, right=0.45cm of modeleval] (resulteval) {Result Evaluation\\Agent 5};

\node[labelstyle, left=1.0cm of datawrangling] {Machine Roles};

\draw[dashedline] ([yshift=-0.50cm]datawrangling.south west |- datawrangling.south)
  -- ([yshift=-0.50cm]resulteval.south east |- resulteval.south);

\node[box, below=1.00cm of datawrangling] (metriceval) {Metric Evaluation\\Human 2 \& Agent 1};
\node[labelstyle, left=1.0cm of metriceval] {Human--Machine\\Collaboration};

\end{tikzpicture}
\end{adjustbox}
\caption{Human--Machine Role Allocation and Oversight Framework}
\label{fig:human_machine_roles}
\end{figure}

\begin{tcolorbox}[
  title={Role Separation and Independent Oversight (RS)},
  colback=white,
  colframe=gray!70,
  fonttitle=\bfseries,
  boxrule=0.6pt,
  arc=1pt,
  left=4pt,
  right=4pt,
  top=2pt,
  bottom=0pt,        
  before skip=4pt,
  after skip=4pt
]

\begin{equation*}
\label{ap_eq:human_governed}
RS(x) \;=\; \frac{\left|\mathcal{E}_{\mathrm{sep}}(x)\right|}{\left|\mathcal{E}^{\star}\right|} \;\in\; [0,1]. 
\end{equation*}

\noindent\textbf{Illustrative Inputs:}
\begin{itemize}\setlength{\itemsep}{0.2em}\setlength{\topsep}{0pt}
 \item $\mathcal{E}_{\mathrm{sep}}(x) = 12$;
 \item $\mathcal{E}^{\star} = 12$.
   
\end{itemize}

\begin{equation*}
\label{ap_eq:human_governed_num2}
\mathrm{RS}(x)
=
\frac{12}{12}
=
1.
\end{equation*}

\end{tcolorbox}

\begin{tcolorbox}
[
  title={Total Self-Served Purpose (S) Score},
  colback=white,
  colframe=black,
  fonttitle=\bfseries,
  boxrule=0.6pt,
  arc=1pt,
  left=4pt,
  right=4pt,
  top=2pt,
  bottom=2pt,
  before skip=4pt,
  after skip=4pt
]

\vspace{0.4em}
\begin{equation*}
\label{ap_eq:intimacy_overall2}
\begin{aligned}
S(x)
&=
w_{\mathrm{TL}}^{S}\,\mathrm{T}(x)
+
w_{\mathrm{RS}}^{S}\,\mathrm{RS}(x).
\end{aligned}
\end{equation*}

\vspace{0.3em}
\noindent\textbf{Illustrative Inputs (Equal Weights and T is not applicable):}
\begin{itemize}
    \item $w_{\mathrm{HG}}^{I} = 1$;
    \item $\mathrm{T}(x)$ is not applicable;
    \item $\mathrm{RS}(x) = 1$ (Role Separation and Independent Oversight);
\end{itemize}

\vspace{0.3em}
\begin{equation*}
\label{ap_eq:intimacy_example2}
S(x)
=
1 \cdot 1
=
1.
\end{equation*}

\end{tcolorbox}

\begin{tcolorbox}[
  title={Human--Machine Trust Formulation},
  colback=white,
  colframe=black,
  fonttitle=\bfseries,
  boxrule=0.6pt,
  arc=1pt,
  left=4pt,
  right=4pt,
  top=2pt,
  bottom=2pt,
  before skip=4pt,
  after skip=4pt
]

\begin{equation}
\label{ap_eq:trust}
\begin{aligned}
\tau(x)
&=
\sigma\!\left(
\alpha_C C(x)
+
\alpha_R R(x)
+
\alpha_I I(x)
-
\alpha_S S(x)
\right),&
\sigma(z)=
\frac{1}{1+\exp(-z)} .
\end{aligned}
\end{equation}

\vspace{0.3em}
\noindent\textbf{Illustrative Inputs (Equal Weights):}
\begin{itemize}\setlength{\itemsep}{0.2em}\setlength{\topsep}{0pt}
    \item $\alpha_C = \alpha_R = \alpha_I = \alpha_S = \tfrac{1}{4}$;
    \item $C(x) = 0.85$ (Credibility);
    \item $R(x) = 0.83$ (Reliability);
    \item $I(x) = 0.32$ (Intimacy);
    \item $S(x) = 1.00$ (Self-Served Purpose).
\end{itemize}

\vspace{0.3em}
\begin{equation*}
\label{ap_eq:trust_num}
\begin{aligned}
\tau(x)
&=
\sigma\!\left(
\tfrac{1}{4}\cdot 0.82
+
\tfrac{1}{4}\cdot 0.83
+
\tfrac{1}{4}\cdot 0.32
-
\tfrac{1}{4}\cdot 1
\right) \\[4pt]
& 
=
\frac{1}{1+\exp(-0.2425)}
\;\approx\; 0.56.
\end{aligned}
\end{equation*}

\end{tcolorbox}

\newpage
\section{SR-Delta}
Based on the dataset extracted from Luxshare and the MSCI 2024 methodological guidelines, we identify two primary categories of discrepancies: ambiguities in disclosure definitions and over- or under-penalization.

\subsection{Disclosure Definition Ambiguity Example}

\newcommand{\boxdivider}{\par\vspace{0.3em}\noindent\rule{\linewidth}{0.6pt}\vspace{0.5em}\par}

\begin{table}[ht]
\centering
\renewcommand{\arraystretch}{1.25}
\setlength{\tabcolsep}{6pt}
\label{tab:stride_example}

\begin{tabularx}{\columnwidth}{|>{\raggedright\arraybackslash}X|}
\hline
\textbf{MSCI Results:} The Executive Pay Disclosure scored -1.2. \\
\hline
\textbf{STRIDE-Guided Extraction Result:} An extracted GRI-aligned metric
"Has the company failed to disclose specific pay totals for its top executives,
including the CEO?". The answer in our dataset is No, which means that -1.2 does not apply\\[0.3em]

\hline

\textbf{Reference Definition from \href{https://www.msci.com/documents/1296102/34424357/MSCI+ESG+Ratings+Methodology+-+Pay+Key+Issue.pdf/5bbf5a6c-cdd4-c7ba-87a0-2f8dc2bf4b96}{MSCI Methodology (Page 9)}:} \\[0.3em]
Has the company failed to disclose specific pay totals for its top executives,
including the CEO? \\[0.3em]
Flagged if yes. \\[0.3em]
This is the most basic test in this area, and as such, carries considerable weight
when invoked, as it must serve in the place of several of the other metrics normally
applied in this area. Disclosure must include, at minimum, pay for all executive
members of the board of directors (or management board for companies with a two-tier
board structure) on an individualized basis or, in cases where there are no executive
members, the CEO. The individualized disclosure should include separate information
on the amount of each of the following (where such a component is provided): salary,
short-term incentives, long-term incentives, pensions, benefits, one-off payments
(such as recruitment or retention awards). \\

\hline

\textbf{Analysis:} There are two plausible interpretations of how the scoring is applied:
\begin{enumerate}
  \item Disclosure of an aggregate total would be considered sufficient, even if
        detailed breakdowns are missing.
  \item Points be deducted whenever information is not provided for each compensation category even the total disclosed.
\end{enumerate} \\

\hline

\textbf{Conclusion:} \\[0.3em]
The company disclosed the total compensation amount. Therefore, the methodologies description can be adjusted to account for potential alternative interpretations. \\

\hline
\end{tabularx}
\end{table}

\newpage
\subsection{Scoring Errors}

\begin{table}[ht]
\centering
\renewcommand{\arraystretch}{1.25}
\setlength{\tabcolsep}{6pt}
\label{tab:stride_chemical_safety}

\begin{tabularx}{\columnwidth}{|>{\raggedright\arraybackslash}X|}
\hline
\textbf{MSCI Results:} Chemical Safety scored 6.1\\
\hline

\textbf{STRIDE-Guided Extraction Result:} \\[0.3em]
Luxshare discloses compliance with major chemical regulations (RoHS, REACH),
internal EHS management systems, supplier chemical audits, and incoming material
inspections. However, the company does not disclose any chemical phase-out roadmap, named priority hazardous substances, quantitative substitution targets, public restricted substance lists (RSL/MRSL), consumer-facing chemical transparency, or time-bound elimination commitments. No supplier non-compliance rates or year-over-year chemical risk reduction metrics are reported. \\
\hline

\textbf{Reference Definition from \href{https://www.msci.com/documents/1296102/34424357/MSCI+ESG+Ratings+Methodology+-+Chemical+Safety+Key+Issue.pdf/266c662c-7d64-b38f-61ee-3387fd789118?t=1666182593419}{MSCI Methodology (Page 7)}:} \\[0.3em]

"MSCI evaluates companies' exposure to and management of chemical safety risks,
including efforts to reduce or eliminate hazardous substances in products and supply chains. Strong management is characterized by robust chemical management systems, supplier controls, substitution strategies, disclosure of restricted substance lists, and proactive phase-out programs aligned with leading frameworks" \\
\hline

\textbf{Analysis:} Luxshare demonstrates baseline regulatory compliance and internal chemical controls, which support a mid-range score. However, the absence of disclosed substitution strategies, named priority chemicals, phase-out timelines, public RSL/MRSL, and quantitative performance indicators indicates that Luxshare does not meet MSCI's criteria for ``advanced'' or ``robust'' chemical management. The current MSCI score (6.1) appears to reward compliance-level practices rather than best-practice leadership, creating a mismatch between disclosed evidence and scoring outcomes. \\
\hline

\textbf{Conclusion:} \\[0.3em]
Based on STRIDE-guided extraction and MSCI's own methodological definitions,
Luxshare's Chemical Safety management should be classified as moderate rather than strong. A more defensible score range would be approximately 5--5.5 instead of 6.1, reflecting regulatory compliance and supplier oversight without advanced phase-out strategies or transparent chemical risk reduction programs. \\
\hline

\end{tabularx}
\end{table}

\section{Call to Action}

Sustainability and climate have a tight schedule. Sustainability performance continues to be recognized as a proxy for long-term resilience and organizational health of a firm. As a result, measurement and rating systems play a critical role in shaping investor perceptions, regulatory oversight, and corporate incentives. Developing a robust and consensus-driven sustainability rating framework is therefore essential to align decision-making with long-term sustainability goals and reduce fragmentation across existing methodologies.

\begin{enumerate}
    \item We call on the community to contribute perspectives and evidence to iteratively refine and strengthen the STRIDE trust equation.
    \item The proposed frameworks have not yet been tested at scale. We invite the research and practitioner community to participate in large-scale empirical experiments to validate, refine, and strengthen these approaches.
    \item Sustainability is a collective-action problem and a rapidly evolving domain. We encourage a broad participation to foster open dialogue for developing comparable and actionable measurement.
    
\end{enumerate}

\end{document}